\documentclass[sigconf]{acmart}
\AtBeginDocument{%
  \providecommand\BibTeX{{%
    \normalfont B\kern-0.5em{\scshape i\kern-0.25em b}\kern-0.8em\TeX}}}

\setcopyright{acmcopyright}
\copyrightyear{2023}
\acmYear{2023}
\acmDOI{XXXXXXX.XXXXXXX}

\acmConference[HRI '23]{Human Robot Interaction}{March 13--16,
  2023}{Stockholm,  SE}
\acmPrice{15.00}
\acmISBN{978-1-4503-XXXX-X/23/06}

\usepackage{amsmath,stmaryrd,mathtools}
\usepackage{xcolor}
\usepackage{bbm}
\usepackage[toc,page]{appendix}
\usepackage{ifthen}
\usepackage[normalem]{ulem}

\newboolean{include-notes}
\setboolean{include-notes}{true}

\newcommand{\edit}[1]{\ifthenelse{\boolean{include-notes}}{\textcolor{black}{{#1}}}{\textcolor{black}{{#1}}}}  

\newcommand{\remove}[1]{\ifthenelse{\boolean{include-notes}}{\textcolor{red}{\sout{#1}}}{}}

\definecolor{orange(sae/ece)}{rgb}{1.0, 0.49, 0.0}

\newcommand{\uH}{u_\mathrm{H}}

\newcommand{\param}{\theta_\mathrm{H}} 
\newcommand{\weights}{\phi} 

\newcommand{\paramdyn}{f_L}
\newcommand{\approxparamdyn}{f^{\weights}_L}

\newcommand{\jointstate}{x}
\newcommand{\jointdyn}{f}

\newcommand{\uR}{u_\mathrm{R}}

\newcommand{\jointstatetraj}{\mathbf{\jointstate}}
\newcommand{\uHtraj}{\mathbf{\uH}}

\newcounter{Hnum} 

\newcommand\showHnum{\stepcounter{Hnum}\theHnum}

\newcommand\figref{Figure~\ref}

\begin{document}

\title{Towards Modeling and Influencing \\the Dynamics of Human Learning}



\author{Ran Tian} 
\authornote{This work supported by ONR YIP, NSF NRI, and WeRide Corp. Author emails: {\tt\{rantian, tomizuka, anca, abajcsy\}@berkeley.edu}. Project website with link to code: \href{https://sites.google.com/berkeley.edu/midle}{https://sites.google.com/berkeley.edu/midle}.}
\affiliation{%
  \institution{UC Berkeley}
  \streetaddress{}
  \city{}
  \country{}}
\author{Masayoshi Tomizuka}
\affiliation{%
  \institution{UC Berkeley}
  \streetaddress{}
  \city{}
  \country{}}
\author{Anca D. Dragan}
\affiliation{%
  \institution{UC Berkeley}
  \streetaddress{}
  \city{}
  \country{}}
\author{Andrea Bajcsy}
\affiliation{%
  \institution{UC Berkeley}
  \streetaddress{}
  \city{}
  \country{}}




\begin{abstract}
Humans have \emph{internal models} of robots (like their physical capabilities), the world (like what will happen next), and their tasks (like a preferred goal). However, human internal models are not always perfect: for example, it is easy to underestimate a robot's inertia. 
Nevertheless, these models \edit{change} and improve over time as humans gather more experience. Interestingly, robot actions \emph{influence} what this experience is, and therefore influence how people's internal models change. \edit{In this work we take a step towards enabling} robots to understand the influence they have, leverage it to better assist people, and \edit{help} human models more quickly align with reality. Our key idea is to model the human's learning as a nonlinear dynamical system which evolves the human's internal model given new observations. We formulate a novel optimization problem to infer the human's learning dynamics from demonstrations that naturally exhibit human learning. We then formalize how robots can influence human learning by embedding the human's learning dynamics model into the robot planning problem. \edit{Although our formulations provide concrete problem statements, they are intractable to solve in full generality. We contribute an approximation that sacrifices the complexity of the human internal models we can represent, but enables robots to learn the nonlinear dynamics of these internal models.} We evaluate our inference and planning methods in a suite of simulated environments and an in-person user study, where a 7DOF robotic arm teaches participants to be better teleoperators. \edit{While influencing human learning remains an open problem, our results demonstrate that this influence is possible and can be helpful in real human-robot interaction.}
\end{abstract}





\maketitle

\section{Introduction}

Imagine 
your first time controlling a robot arm to perform daily living tasks like throwing away trash 
or stirring a pot of soup. 
Initial interactions with the robot are tough: you aren't familiar with the robot's dynamics so your motions are jerky and imprecise. 
In other words, your \textit{internal model} of the robot is incorrect. 
And robot dynamics are not the only thing we, humans, have incorrect internal models of. 
We might not fully understand the 
world's dynamics (e.g., \edit{result of pouring} lemon juice into cream) or our own preferences
(e.g., \edit{only liking something after trying it}). 

\begin{figure}[t!]
    \centering
    \includegraphics[width=0.44\textwidth]{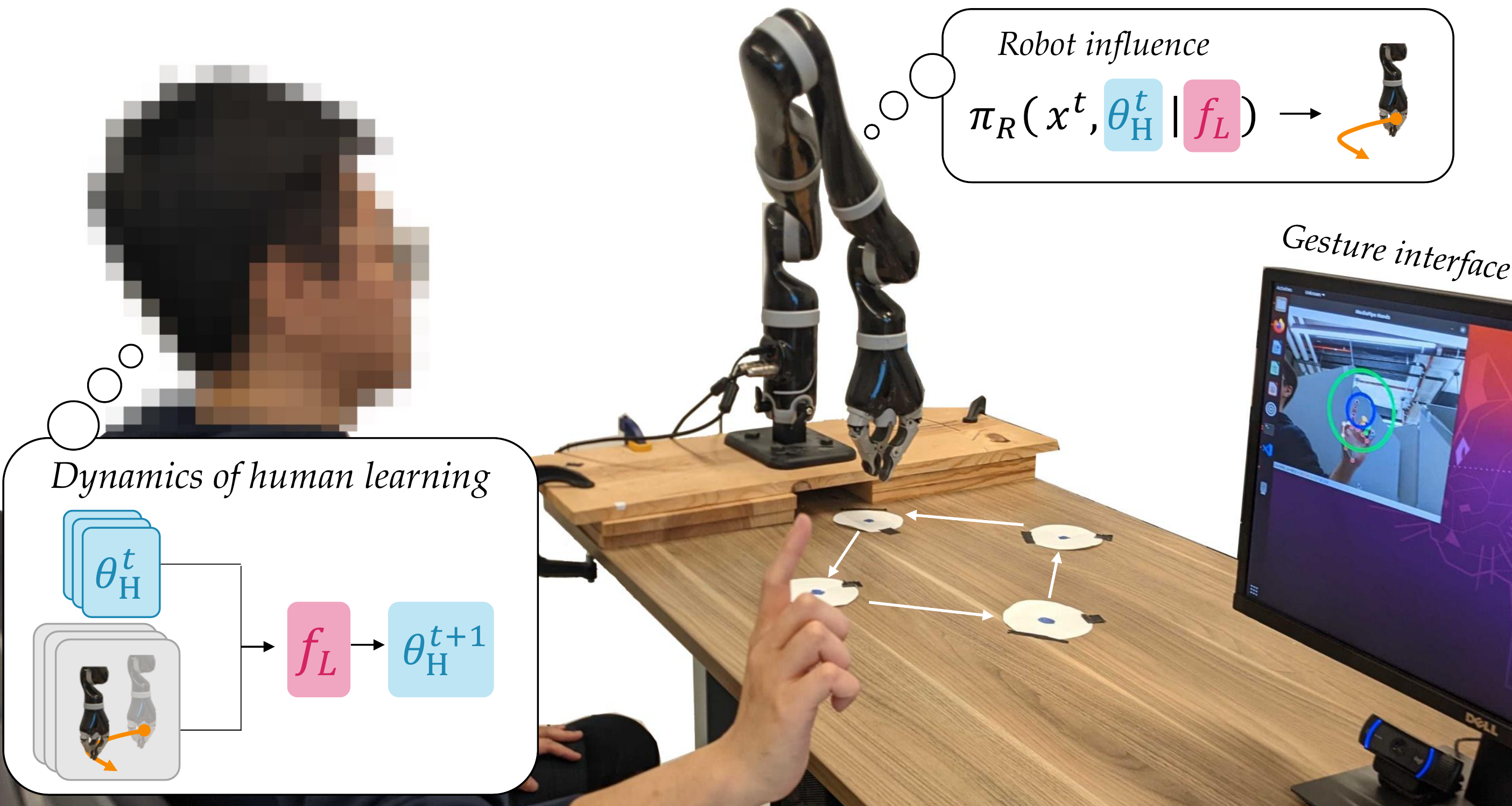}
    \vspace{-0.5em}
    \caption{Human teleoperates a new robot; \textcolor{black}{they update their internal model by acting and observing outcomes}. Planning with human learning dynamics, the robot influences the human's internal model \textcolor{black}{to help} them \textcolor{black}{be} a better teleoperator.}
    \vspace{-2em}
    \label{fig:front_fig}
\end{figure}

However, over time, our internal models evolve \edit{with our experiences}. As you control the robot, you start to understand how it will move; as you try different things, you learn what you like. 
Since the robot is part of the world, the robot's actions and their outcomes 
become part of these experiences. In other words, robot actions inevitably \emph{influence} the change in a human's internal model.

In this work, we advocate that robots should understand and use this influence. First, collaborative tasks require understanding what people are trying to do in order to assist them. Prior work has shown that inferring a human's internal model is critical for assistance \edit{\cite{reddy2018you}}; in turn, we argue that if this model changes over time, tracking this change will enhance assistance. 
Further, purposefully influencing a change in the human's internal model opens the door for \emph{teaching}: robot actions that are optimized to quickly align the human's understanding with reality. For instance, as an operator controlling the robot (\figref{fig:front_fig}), this means you quickly understand how the robot works and can do the task independently. 


A key challenge towards this is \edit{modeling} \textit{how} humans learn; without a proper model of this, the robot cannot plan to change the human's internal model. 
Although we do not know the precise functional form of how people learn,
we observe that a human's understanding of the robot or the world changes as a function of what they observe.
For example, at first you may mistakenly believe that the robot doesn't experience any inertia. 
However, as soon as you gesture to move the robot forward, you see the robot lagging behind. 
This observation controls the evolution of your internal robot physics model. 
The same holds true for your internal model of the world and personal preferences. 
In other words, we can model
\begin{quote}
    \textit{human learning as a dynamical system where the human's internal model is the state, and \edit{the observations---which the robot can influence---evolve the internal model}.}
\end{quote}
\edit{Of course, this does not prescribe the functional form of the dynamical system. One idea is to draw on computational cognitive science work to define this function. A predominant lens is that of probabilistic models \cite{goodman2016probabilistic}, which posits that humans perform some form of approximate Bayesian inference based on the observations they receive. 
In reality, people have been shown to have a plethora of cognitive biases which deviate from perfect Bayesian inference: they might use gradient information \cite{weiss2002motion}, might not process the entire observation due to sensory overload \cite{mulder1999cybernetics}, or exhibit systematic bias like over- or under- estimation \cite{reddy2020assisted}.  
Instead of committing to a specific model, in our work we treat this as a general dynamics learning problem, which has roots in controls and robotics \cite{khalil2002modeling}. We leverage demonstrations which \textit{naturally} exhibit human learning (e.g., humans teleoperating a robot they have never interacted with before), to fit a human learning model under the assumption that observed human actions are approximately optimal given their current internal model. This enables flexibility of capturing different possible learning updates, at the cost of being domain-specific.}

Although \edit{the most general model learning problem remains computationally intractable},
we introduce a tractable approximation that is readily solvable via gradient-based optimization, and is compatible with neural network representations of the human learning dynamics. 
Leveraging our approximate dynamics model of human learning, 
we formalize robot influence over the human's internal model as a Markov Decision Process (MDP) where the human's internal model is part of the state and the human's learning dynamics are part of the transition function. 
The solution yields robot actions that change the human's internal model by changing the human's observations in a way that rewards the robot. 

We run experiments with simulated humans to study the fidelity of the inferred human learning dynamics and investigate robot teaching and assistance in settings where the human's understanding of robot physics, motion preferences, or goals can be influenced. 
Finally, we conduct a user study with a Kinova Jaco 7DOF robot arm and find \edit{that our method can help} teach humans to be better teleoperators. \edit{Overall, while influencing human learning remains an open problem, we are excited to have taken a step in this domain via a principled yet tractable learning and planning method.}






\section{Related Work}

\noindent 
\textbf{Inferring human preferences and beliefs.}
A large body of work has focused on 
learning human reward functions via inverse reinforcement learning (IRL) \cite{ng2000algorithms, jara2019theory, kitani2012activity}. 
This includes inferring human driving preferences \cite{sadigh2016planning, peters2021inferring}, desired exoskeleton gaits \cite{li2021roial}, intended goals \cite{jain2019probabilistic}, motion preferences \cite{pfeiffer2016predicting}, \edit{and human understanding about physics \cite{reddy2018you}}. 
A key assumption in these works is that people have \textit{static} internal models of preferences or physics.
Instead, we are interested in learning a \textit{dynamic} model of how humans change their preferences, goals, and understanding of physics. 

\smallskip
\noindent \textbf{Models of human learning \edit{for robot decision-making}.}
Prior works \edit{in robotics} model human learning as Bayesian inference when updating goals or preferences \cite{dragan2013legibility, huang2019enabling, habibian2022encouraging}, a linear Gaussian system when updating trust \cite{chen2020trust}, gradient-based IRL when learning rewards \cite{cakmak2012algorithmic}, or as 
a multi-armed bandit algorithm when updating preferences \cite{chan2019assistive}.
\edit{Instead of assuming a known model of how people learn,}
in this work we seek to \textit{learn} a model of how humans learn.
Most related to our work is \cite{reddy2020assisted} which learns a model of how people estimate the state of the world. 
In this work, we propose a generalization where the human is not estimating world state, but updating their preferences, goals, and internal physics model. This induces a significantly harder model learning problem, for which we propose a tractable approximation.  

\smallskip
\noindent \edit{\textbf{Cognitive theories of human learning.} 
Models of human inference have been extensively studied in both computational cognitive science \cite{griffiths2010probabilistic, baker2009action} and psychology \cite{premack1978does, annurev-devpsych-121318-084833}. 
While human cognition can be broadly modeled at three levels (computational, algorithmic, and hardware) \cite{marr2010vision}, most relevant to us are the algorithmic works.
\cite{griffiths2010probabilistic} posits that 
modeling human reasoning as ``implementing'' an exact Bayesian posterior or a gradient-based point estimate are both compatible with probabilistic models of human cognition, and are a potential source of rational process models \cite{shi2008performing}. 
Further, \cite{schaefer2012beside} finds evidence that humans may update their forward models using the models' prediction error as loss functions. 
Inspired by these works, our simulated human experiments leverage exact and approximate probabilistic inference models, and we study if our flexible, learning-based method can effectively recover such models. } 

\smallskip 
\noindent \textbf{Robot influencing human behavior.}
While there are many ways a robot can influence humans (e.g., through nonverbal cues, appearance, visuals, or curriculum design \cite{saunderson2019robots, rae2013influence, admoni2017social, reddy2020assisted, srivastava2022assistive}), we focus on robot influence through physical action \cite{newman2020examining}.
A common approach towards this
models human-robot interaction as a game \cite{sadigh2016planning, nikolaidis2017game, schwarting2019social, laine2021multi, tian2022safety, hu2022active}. 
While these approaches can capture reactions from the human, they do not address the internal learning problem: over repeated interactions, the human may not have learned anything and is only reacting.
Alternatively, model-free methods learn a latent representation of the human's policy and then leverage the latent dynamics to influence the human \cite{xie2020learning, parekh2022rili}.
Here the human's internal model is \textit{implicitly} captured by the latent representation, and the internal model evolves between interaction episodes.
In contrast, in our work the human's internal model is an \textit{explicit} parameterization (e.g., high-dimensional parameterization like dynamics) and the human internal model can evolve continuously during an interaction episode. 
This enables robot behaviors like teaching the human the correct internal model, which would otherwise not be possible with implicit, latent representations.

\section{Modeling How Humans Learn \& Act}


We begin by \edit{mathematically modelling} the dynamics of human learning, before diving into how the robot can infer this dynamics model and use it \edit{influence the human's internal model evolution.}

\smallskip  
\noindent \textbf{Notation.}
Let $\jointstate \in \mathbb{R}^{n}$ be the state of the world including the robot (e.g., robot end-effector position, objects, etc.). 
Both the human and robot can take actions,  $\uH \in \mathbb{R}^{m}$ and $\uR \in \mathbb{R}^{m}$ respectively, that 
affect the next state. 
Let the deterministic world dynamics be 
\begin{equation}
\jointstate^{t+1} = \jointdyn(\jointstate^t, \uH^t, \uR^t).
\label{eq:physical_dyns}
\end{equation}

\smallskip 
\noindent \textbf{Human internal model.}
We model the human as having an internal parameter vector, $\param$, which captures a latent aspect of the task that the human is uncertain about but \textit{continuously learns about}. 
Going back to our motivating example where the human teleoperates a robot, $\param$ can model the human's current estimate of the robot's physical properties, like its inertia. 
Or, $\param$ could model the human's current preferences for teleoperation: they start off wanting to move the robot to one goal, but then change their mind to a new goal after realizing it is easier to reach. Regardless of what $\param$ represents, it is important to remember that it is \textit{time-varying} and that it \emph{evolves as a function of what the human observes}. 

\smallskip
\noindent \textbf{Human policy: acting under the internal model.}
In our work, we model the human actions as driven by some reward function, $R_\mathrm{H}(\jointstate, \uH; \param)$, which depends on the current state, the human's action, and their internal parameter $\param$. Following prior works \cite{ziebart2008maximum, waugh2010inverse, levine2012continuous, baker2009action}, we treat the human as a noisily-optimal actor:
\begin{align} 
     & \mathbb{P}(\uH \mid  \jointstate; \param) = e^{Q_\mathrm{H}(\jointstate, \uH;\param)}\Big(\int_{\Tilde{u}} e^{Q_\mathrm{H}(\jointstate, \Tilde{u};\param)}d\Tilde{u}\Big)^{-1}, 
    \label{eq:human_policy}
\end{align}
where the optimal state-action value is denoted by $Q_\mathrm{H}(\jointstate, \uH; \param)$ and $\jointstate$ is the current state, $\uH$ is the human action, and $\param$ the human's current parameter estimate. 

We make two simplifying assumptions in this model. First, the human does not explicitly account for the actions $\uR$ the robot could take.
Instead, the human reacts to the current state $\jointstate$, which \textit{implicitly} captures the effect of any robot actions that change the state. 
This models scenarios where the human is doing the task on their own, or where the human is not aware of how the robot is providing guidance. 
Second, when the human \edit{plans} their action, we assume that they separate the estimation of $\param$ from policy generation and \edit{they} plan with their current estimate. 


\smallskip
\noindent \textbf{Dynamics of human learning: updating the internal model.}
As the human acts in the environment, they receive new observations: they may see the next state, including that of the robot's, or experience how much they enjoy something (i.e. observe ``reward signal''). 
This naturally lets the human update their understanding of the robot, physical aspects of the world, or their preferences.

Leveraging our \edit{core} idea, we model the human's learning process as a \edit{nonlinear} dynamical system over the human's internal model parameter. 
Let $\param^0$ be the human's initial internal model, and $x^{0:t}$ and $\uH^{0:t}$ be the state and action history until timestep $t$ and $x^{t+1}$ be the resulting state at the next timestep, possibly including the influence of robot actions. 
Given the initial parameter estimate, the state and action history, and next state data, the human evolves their internal model to the next estimate, $\param^{t+1}$. 
Let the true dynamics of the human's learning process be:
\begin{equation}
    \param^{t+1} = \paramdyn(\param^0, \jointstate^{0:t+1}, \uH^{0:t}).
    \label{eq:human_learn_dyns}
\end{equation}
Here we are faced with the question \textit{``What $\paramdyn$ models how the human learns?''}
Instead of \edit{committing to a specific model, here we take a robotics perspective and view this question as an instance of a dynamics learning problem. By looking to human data, we aim to \textit{learn} an approximate $\paramdyn$ model that is domain-specific}.



\section{Inferring the Dynamics of Human Learning}

In this section we focus on inferring the dynamics of human learning \edit{by leveraging} demonstrations which \textit{naturally} exhibit human learning: for example, initial trials of a human teleoperating a robot they have never interacted with before. We assume these demonstrations contain only the state and action histories and do not contain ground-truth human internal model data (since this is not possible in practice). However, we do assume that the observed actions are coupled with the human's internal model, allowing us to leverage demonstrations to infer the dynamics of the human's internal model.
Given this dataset, we seek to fit a nonlinear model to represent the dynamics of human learning,
\begin{equation}
    \paramdyn^{\phi} \approx \paramdyn,
\end{equation}

\noindent where $\phi$ are the parameters of the approximate model. 
In the following sections, we formalize inferring $\paramdyn^\phi$ as a maximum likelihood estimation (MLE) problem and propose a tractable approximation. 

\subsection{Formalizing the Inference Problem} 
Let $\mathcal{D}_{demo} := \{(\jointstatetraj, \uHtraj)_i\}^N_{i=0}$ be a collection of $N$ demonstrations containing state and human action trajectories of length $T$ time steps. 
We want to infer the parameter of the human's learning dynamics, $\phi$, and the initial human parameter estimate, $\param^0$, which maximizes the likelihood of the observed demonstrations.
We formulate this inference via the constrained optimization problem:
\begin{align}
\max_{\weights,\param^0} \quad & \sum_{(\jointstatetraj, \uHtraj) \in \mathcal{D}_{demo}} \sum_{t=0}^{T-1} \log \Big[\mathbb{P}(\uH^t \mid  \jointstate^t;\param^t) \Big],\label{eq:learning_problem} \\ 
  \textrm{s.t.} \quad & \param^{t+1} = \approxparamdyn(\param^0, \jointstate^{0:t+1}, \uH^{0:t}),
  \label{eq:inferred_learning_dyns}
\end{align}
where $\mathbb{P}(\uH^t \mid  \jointstate^t,\theta^t)$ is the human action likelihood from Equation~\eqref{eq:human_policy} and the constraint ensures that the human's internal parameter evolves according to the human's learning dynamics model.


\subsection{Solving the Inference Problem} 
Unfortunately, the inference problem in Equation~\eqref{eq:learning_problem} is intractable to solve directly for two main reasons. 
First, recall that the human's internal model $\param$ of their preferences, dynamics, or goals, changes over time.
This means that at each timestep the human is generating data $\uH$ under a possibly different $\param$. 
In other words, the human acts under a \textit{new} action policy $\mathbb{P}(\uH^t \mid  \jointstate^t;\param^t)$ at each $t$, requiring us to  
solve an entirely \textit{new} reinforcement learning problem to obtain the action policy at each time step along the inference horizon.
In the case where $\param$ is a continuous, high-dimensional parameter (e.g., physical properties of the robot dynamics), this is intractable to compute per-timestep. 
Secondly, even if we could obtain the human's policy infinitely fast, our optimization problem still requires searching over the the high-dimensional space of $\phi$ and $\param$.
Gradient-based optimization is a natural choice, but we need to be able to compute the gradient of the MLE objective and, therefore, differentiate through $Q_\mathrm{H}$ with respect to $\param$. 


In the following subsections, we introduce several approximations to arrive at a tractable solution to the inference problem. 
Our key idea is to use a linear-quadratic (LQ) approximation of \edit{the physical dynamics and the human reward}. This enables us to derive a closed-form expression of the human policy as a function of $\param^t$ at any time and yields a differentiable inference objective. 


\subsubsection{Linear-Quadratic approximation.}
We take inspiration from infinite-horizon linear-quadratic (LQ) control \cite{kalman1960contributions} and 
assume that the human's reward is quadratic and their model of the physical dynamics is linear.
Let the linear physical dynamics be:
\begin{align}
    \jointstate^{t+1} = \jointdyn(\jointstate^t, \uH^t, \uR^t \equiv 0) &\approx A\jointstate^t + B\uH^t 
    \label{eq:linear_phys_dyns}
\end{align}
where $A \in \mathbb{R}^{n\times n}, B \in \mathbb{R}^{n\times m}$ are matrices governing the physical dynamics. Note that in the human's mind, the robot is not exerting any control effort, and hence $\uR \equiv 0$. 
Let the human's reward be approximated by a quadratic function:
\begin{align}
    r_\mathrm{H}(\jointstate, \uH; \param) \approx -\jointstate^\top Q \jointstate - \uH^\top R \uH,
    \label{eq:human_reward_quadratic} 
\end{align}
where the matricies $Q \in \mathbb{R}^{n \times n}$ and $R \in \mathbb{R}^{m \times m}$ tradeoff the state reward (e.g., how much reward the human gets for reaching a state) and the action reward (e.g., how much effort the human wants to exert), respectively. 
Note that $\param$ enters in different ways depending on what the human is learning about. 
For example, if $\param$ encodes \textbf{reward weights} (i.e., the human's preferences about how to do a task), then $\param := (Q, R)$. 
If the parameter encodes a human's \textbf{goal state}, then $\param \in \Theta \subset \mathbb{R}^n$ and the human's reward function regulates the human towards their desired goal: $r_\mathrm{H}(\jointstate, \uH; \param) \approx -(\jointstate- \param)^\top Q (\jointstate - \param) - \uH^\top R \uH$. 
Finally, if $\param$ encodes aspects of the \textbf{physical dynamics} that the human is estimating, then $\param := (A, B)$ from the dynamics in Equation \eqref{eq:linear_phys_dyns}, and governs how the human imagines the physical dynamics evolving.

\subsubsection{Closed-form $Q_\mathrm{H}$.}
Recall that the human plans a policy using their current estimate $\param$; at every step, $\param$ changes, resulting in a new policy. 
In general, obtaining the exact $Q_\mathrm{H}$-value via dynamic programming in continuous state, action, and $\param$-spaces is computationally demanding. 
However, under our infinite-horizon LQ-approximation 
the human's $Q_\mathrm{H}$-value is:
\begin{align}
    Q_\mathrm{H}(\jointstate, \uH; \param) = r_\mathrm{H}(\jointstate, \uH; \param) -(\jointstate')^\top P_{\param} (\jointstate') \label{eq:q_value_definition}
\end{align} 
where the instantaneous reward is quadratic from Equation~\eqref{eq:human_reward_quadratic} and $\jointstate'$ is the next physical state as a result of applying $\uH$ from state $\jointstate$. 
Note that  $-(\jointstate')^\top P_{\param} (\jointstate')$ is the infinite-horizon optimal value
where $P_{\param}$ is the well-known positive-definite fixed point of the discrete-time algebraic Riccati equation (DARE) \cite{bertsekas2011dynamic}:
\begin{align}
P = A^{\top}PA-A^{\top}PB(R+B^{\top}PB)^{-1}B^{\top}PA+Q.
\label{eq:DARE}
\end{align}
Obtaining $P_{\param}$ also yields the optimal human action: $\uH^*(\jointstate; \param) = -K_{\param} \jointstate$ where $K_{\param} = (R + B^\top P_{\param} B)^{-1}B^\top P_{\param} A$. 
Note that in all of the equations above, $\param$ enters differently depending on what \edit{the} human's internal model represents.

\subsubsection{Closed-form human policy.}
In general, obtaining the human policy in Equation~\eqref{eq:human_policy} is computationally intractable
in continuous action spaces due to the integral over $\uH$.  
However, plugging in our closed-form $Q_\mathrm{H}$, we see that the exponent is quadratic in $u$, allowing us to take a Gaussian integral \cite{tierney1986accurate}.
Overall, this yields a closed-form human policy (\edit{see full derivation in Appendix~\ref{app:gaussian_integral}.}):
\begin{equation}
    \mathbb{P}(\uH \mid \jointstate; \param) = |\mathbf{H}|^{1/2}(2\pi)^{-m_\mathrm{H}/2} e^{Q_\mathrm{H}(\jointstate, \uH;\param) - Q_\mathrm{H}(\jointstate, u^*;\param)}. 
\end{equation}



\subsubsection{Representing the dynamics of human learning}
\edit{Finally, we are faced with the question of how to functionally represent the dynamics of human learning; for example, we could take inspiration from computational cognitive science and model $\approxparamdyn$ as Bayesian inference \cite{goodman2016probabilistic}. 
Instead of committing to a specific functional form, in this work we seek a model that \edit{has the potential to} capture a broad range of ``learning \edit{algorithms}'' that the human could use to update their internal parameter.}
Recently, self-attention based transformer models \cite{vaswani2017attention} have shown success at predicting high-dimensional sequential tasks \cite{janner2021offline}, \edit{at the cost of being domain-specific}. 
Inspired by this, we represent $\paramdyn^\phi$ as a transformer encoder 
where $\phi$ are the weights of the neural network. 
At each time step $t$, a collection of the state $\jointstate^t$, the human's action $\uH^t$, and the next state $\jointstate^{t+1}$ are fed into an encoder to extract embeddings \edit{which are fed into a transformer encoder that} predicts the human's next internal model.
Training details are in Appendix~\ref{app:training_details}. 

\subsubsection{Deriving an efficient, gradient-based solution}
To optimize the transformer-based model of human learning dynamics, we need the gradient of our inference objective with respect to the neural network parameters. \edit{Here a key challenge lies in the human's policy gradient because it requires differentiating through the DARE function, which is non-obvious. However, we leverage recent work \cite{east2020infinite} to obtain the relevant closed-form Jacobians, enabling us to efficiently infer the parameters of $\approxparamdyn$ via gradient-based optimization. More details on this approach are in Appendix~\ref{app:gradient_based_soln}.}

\section{Influencing human learning with robot actions}

Inferring how humans learn presents an opportunity for human-robot interaction.
For example, when a human teleoperator is mistaken about the robot's inertia, it may take them many interactions to learn and become \edit{better}.
Instead, could the robot \textit{influence} the human so that their understanding improves faster?
\edit{Here, we mathematically formalize this influence by embedding the approximate dynamics model of human learning into robot planning}.



\smallskip
\noindent\textbf{Formalizing the Influence Problem.} 
We formalize the robot influence problem as a Markov Decision Process (MDP) where the human's internal model parameter is part of the state.
Our MDP is a tuple $<S, U_\mathrm{R}, T, r_\mathrm{R}>$ where the state $s = (\jointstate, \param) \in S$ is the joint physical state and human internal model parameter and the robot's actions are $\uR \in U_\mathrm{R}$. The stochastic state transition function is defined as
$T(s^{t+1} \mid s^{t}, \uR^t) := \sum_{\uH} \mathbb{P}(\uH \mid s^t) \tilde{f}(s^{t}, \uR^t, \uH^t, s^{t+1})$
which accounts for the human policy from Equation~\eqref{eq:human_policy}.
Importantly, $\tilde{f}(s^{t}, \uR^t, \uH^t, s^{t+1})$ is a deterministic function that evolves $x^t$ via the physical dynamics $f$ from Equation~\eqref{eq:physical_dyns} and the human's internal model parameter $\param^t$ via the human learning dynamics $\approxparamdyn$ from Equation~\eqref{eq:inferred_learning_dyns}. 
Finally, the robot optimizes its reward function $r_\mathrm{R}(s, \uR, \uH; \theta^*)$ where $\theta^*$ is the robot's \textit{true} internal model parameters (e.g., the robot's true physical dynamics). 
Note that because $s = (\jointstate, \param)$, the robot's reward depends on the human's time-varying internal model, $\param$, at each timestep. 

The robot seeks an optimal policy $\pi^*_\mathrm{R}$ which maximizes it's reward in expectation over the human's action sequence, $\uHtraj$:
\small
\begin{equation}
\begin{aligned}
\pi^*_\mathrm{R} = \arg\max_{\pi_\mathrm{R}}  \mathbb{E}_{\uHtraj}\Big[ \sum_{t=0}^{\infty} r_\mathrm{R}(s^t, \uR^t, \uH^t; \theta^*)\Big] ~~\textrm{s.t.} ~~ T(s^{t+1} \mid s^{t}, \uR^t),
\end{aligned}
\label{eq:robot_planning_problem}
\end{equation}
\normalsize
Because human's internal model parameter $\param^t$ is part of the state and the state transition function $T(s^{t+1} \mid s^{t}, \uR^t)$ includes the inferred dynamics model of human learning, $\pi_R^*$ should automatically influence the human's internal model if it yields higher reward.

\smallskip
\noindent\textbf{Computing Solutions to the Influence Problem}
The presence of the human's nonlinear learning dynamics $\approxparamdyn$ in the transition function results in a nonconvex optimization problem. 
To obtain the optimal robot policy, we would have to solve the MDP either exactly with dynamic programming (which suffers from the curse of dimensionality) \cite{bertsekas2011dynamic} or approximately via receding-horizon control (which requires trading off optimality with computational efficiency) \cite{camacho2013model}.
To achieve both long-horizon reasoning and efficient runtime performance, we use a Dyna-style algorithm \cite{sutton2018reinforcement} that uses the samples generated by the transition $T(s^{t+1} \mid s^{t}, \uR^t)$ to train $\pi^*_R$ using model-free learning (Proximal Policy Optimization \cite{schulman2017proximal}).


\section{Simulated Human Experiments}

We want to test two aspects of our approach: our ability to infer the dynamics of human learning and the effectiveness of our robot influencing algorithm.
To fully validate both, we need access to the ground-truth human learning dynamics ($\paramdyn$). 
For this reason, we first perform a series of simulation experiments with simulated humans. We explore two shared autonomy contexts: a robot teaching a human about physics-based robot dynamics (Section \ref{subsec:teaching_physics_dyn}) and a robot that implicitly influences human objectives, like their goal or motion preferences (Section \ref{subsec:reward_manipulation}).

Similar to prior work in shared autonomy \cite{dragan2013policy, jain2019probabilistic, newman2022harmonic, losey2022learning}, the robot combines the human's commanded action, $\uH$, with the robot's planned guidance, $\uR$, and executes the action:  
\begin{equation}
    u = \alpha \cdot \uR + (1 - \alpha) \cdot \uH
    \label{eq:shared_control}
\end{equation} 
where $\alpha \in [0,1]$ trades off how much guidance the robot can exert. 
In all experiments, we use $\alpha = 0.5$. 
To generate human demonstrations and infer the human learning dynamics, we simulate a suite of human learners (see \ref{subsubsec:sim_humans_phys} and \ref{subsubsec:sim_humans_reward}). 
In each experimental environment we collect 50 demonstrations for model learning. We randomize the initial state of the robot for each demonstration, and randomize the robot actions during each interaction.\footnote{We randomize $\uR$ to diversely cover how human's internal model changes.}


\begin{figure}[h!]
    \centering
    \includegraphics[width=0.45\textwidth]{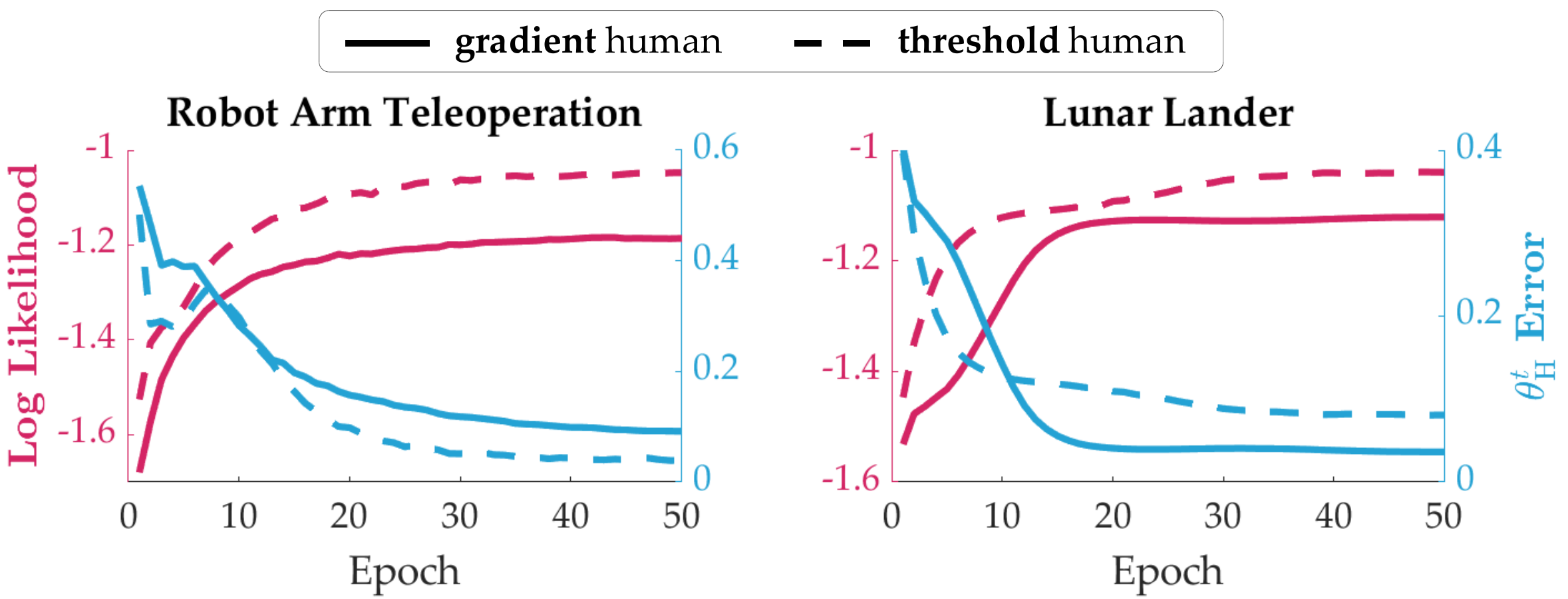}
    \vspace{-1em}
    \caption{Our inference problem lets us learn to predict $\param^t$.}
    \label{fig:dyn_model_learning}
    \vspace{-1em}
\end{figure}

\subsection{Teaching Physical Dynamics}
\label{subsec:teaching_physics_dyn}
We focus on shared autonomy \edit{settings} where the human knows the task objective (e.g., control a robot arm to follow a path), but they learn about the true robot dynamics (e.g., inertia).
We want to understand how the human learns about the physical robot dynamics, and if a robot that actively \textit{teaches} the human about its \edit{physics} can help the human quickly improve their task performance. 

\begin{figure*}[t!]
    \centering
    \includegraphics[width=0.92\textwidth]{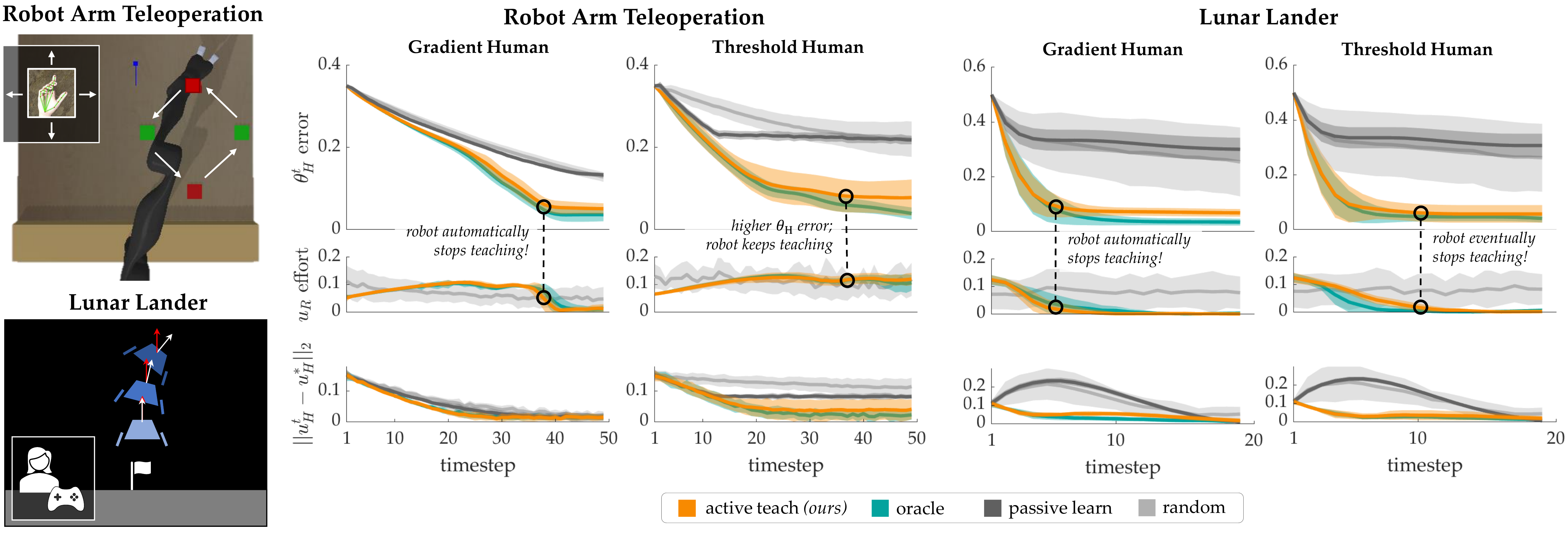}
    \vspace{-1.2em}
    \caption{(left) Visualization of both simulation environments. (right) Mean and standard deviation of human internal model error, robot effort, and human action optimality for both dynamics teaching environments, and both simulated humans.}
    \label{fig:teleop_dyn_results}
    \vspace{-1em}
\end{figure*}

\subsubsection{Dynamics of human learning.}
\label{subsubsec:sim_humans_phys}
\edit{Motivated by computational cognitive science models \cite{griffiths2010probabilistic},}
we simulated two types of human learners: \textbf{gradient-based} learners and \textbf{threshold} learners. 
All humans update their internal model via Equation~\eqref{eq:human_learn_dyns}, but the structure of $\paramdyn$ takes various forms.
After observing a new state-action pair $(\jointstate^t, u^t)$, the \textbf{gradient-based} learner updates their parameter $\param^t$ according to a gradient-ascent update rule: 
$\paramdyn^{\textrm{grad}} := \param^{t} + \eta \nabla_{\param} P(u^t \mid \jointstate^t; \param^t)$
where $\eta \in \mathbb{R}_{+}$ is the step size. 
Note that $u^t$ is the observed, total executed control, possibly combining $\uR$ and $\uH$.
Intuitively, this learner can be viewed as doing gradient-based maximum likelihood estimation of their latent parameter, similarly to prior IRL methods \cite{ziebart2008maximum}. 
The \textbf{threshold} learner also uses a gradient-based learning rule, but only updates their internal parameters if they observe a ``large enough'' change: $\paramdyn^{\textrm{thresh}} := \param^{t}  + \eta \mathbbm{1}_{|\nabla P(u^t \mid \jointstate^t, \param^t)| > \epsilon} \big[\nabla_{\param} P(u^t \mid \jointstate^t, \param^t) \big]$
where $\mathbbm{1}$ is an indicator determining if the magnitude of the gradient is deemed large enough to induce a learning update and $\epsilon$ is a threshold parameter. 
 

\subsubsection{Human internal model.}
In all experiments, the simulated humans are learning about the robot's physical dynamics and thus $\param$ encodes various aspects from Equation~\eqref{eq:linear_phys_dyns}. 

\subsubsection{Simulated environments.} \label{subsubsec:phys_dyns_sim_envs}
\figref{fig:teleop_dyn_results} shows our simulated environments, all or which have continuous state and action spaces. 

\smallskip
\noindent \textbf{(1) Lunar Lander.}
The human controls the Lunar Lander's engines to change its tilt. The human wants to keep the lander upright during its descent. 
Let the state be the tilt angle with respect to the ground and tilt angular velocity $\jointstate = (\psi, \omega)$ and $u$ be the engine force.
The dynamics are $\jointstate^{t+1} = A\jointstate^{t} + Bu^t$ where the ground-truth dynamics are $A^* = [1, 0.2; 0, 1], B^* = [0; 0.5]$. Here, the human's internal model represents the control matrix $\param := B$, which depends on the human's inertia estimate.  

\smallskip
\noindent \textbf{(2) Robot Arm Teleoperation.} 
The human controls the end-effector of a 7DOF robot arm via hand gestures (see \figref{fig:teleop_dyn_results}). 
They want to control the robot to reach a series of known goals, $x_g$.
However, one of the robot motors is slightly defective, causing the robot to consistently lag in one direction. 
Let the state be the robot end-effector position $x = (p^x, p^y, p^z)$ and the control $u$ be linear velocity. 
The robot's end-effector dynamics can be described by the goal-dependent system\footnote{Although this system is nonlinear, since the robot knows the human's goal at each time step, the dynamics can be approximated by a linear system $\jointstate^{t+1} = A\jointstate^t + B\Tilde{u}^t $, where $\jointstate^0$ is the system state at that time step and $ \Tilde{u}^t:= u^t - \mathrm{sign}(\jointstate^0 - x_g) \odot w$. }
: $\jointstate^{t+1} = A\jointstate^t + B\big[u^t - \mathrm{sign}(\jointstate^t - x_g) \odot w\big] $
where $w$ is the bias induced by the defective robot motor and $\odot$ is the Hadamard product.
Intuitively, this describes a dynamical system that consistently experiences lag in the $x$-direction. The ground-truth dynamics are $A^* = I^{3 \times 3}$, 
$B^* = \mathrm{diag}(0.4, 0.4, 0.4)$, 
and $w^* = [-0.15, 0, 0]^\top$. The human's internal model is $\param := (B, w)$, which captures their system responsiveness and bias estimates.



\subsubsection{Human objective.}
We assume the human always knows the objective, and their reward function is quadratic as in \eqref{eq:human_reward_quadratic}. For \textbf{Lunar Lander} the human was rewarded for keeping the lander upright and stable ($\psi = 0$, $\omega = 0$), and for \textbf{Robot Arm} they were rewarded for reaching all the goals and tracking the path shown in \figref{fig:teleop_dyn_results}.

\subsubsection{Robot objective.}
The robot objective is to align human's internal model with the true robot dynamics model while minimally intervening. Mathematically, the robot's reward function is:
\begin{equation}
    r_\mathrm{R}(s, \uR, \uH; \theta^*) = -||\param - \theta^*||_2^2 - ||u - \uH||^2_2,
\end{equation}
where the true dynamics are $\theta^* := B^*$ in the \textbf{Lunar Lander} environment and $\theta^* := (B^*, w^*)$ in the \textbf{Robot Arm} setting.

\subsubsection{Baselines.}
We compare our method where the robot actively teaches by planning with the inferred learning dynamics $\approxparamdyn$ (\textbf{Active Teach}) to a robot that teaches with the true learning dynamics $\paramdyn$ (\textbf{Oracle}), no robot intervention (\textbf{Passive Learn}), and a robot that randomly perturbs the human actions (\textbf{Random}).

\subsubsection{Hypotheses.}
\textbf{H\showHnum:} \textit{We can learn to predict $\param^t$ well by maximizing the MLE objective}. \textbf{H\showHnum:} \textit{\textbf{Active Teach} outperforms \textbf{Passive Learn} and \textbf{Random} in aligning the human's internal model}. \textbf{H\showHnum:} \textit{Robot stops intervening when the human's internal model is well-aligned}.

\subsubsection{Results.} For $\mathbf{H1}$, we study the relationship between the MLE objective in \eqref{eq:learning_problem} and our inferred model's ($\approxparamdyn$) ability to predict $\param$. 
\figref{fig:dyn_model_learning} shows these curves for both the \textbf{Robot Arm} and \textbf{Lunar Lander} environments over 50 epochs. 
We see that across both \textbf{gradient} and \textbf{threshold} human learners, the log likelihood of the human's actions increases (shown in pink) while the $\param$ prediction error decreases (shown in blue), supporting $\mathbf{H1}$. 

\figref{fig:teleop_dyn_results} shows the human's internal model error, the robot's effort, and the difference between the human's action and the optimal action in the \textbf{Robot Arm Teleoperation} and the \textbf{Lunar Lander} environment for both types of human learners. We see that across all environments, our method performs comparably to \textbf{Oracle} model, and is able to align the human's internal model of the robot's dynamics with the true dynamics significantly faster than \textbf{Passive Learn} or \textbf{Random} (supporting \textbf{H2}). 
Interestingly, in all but one setting does the robot automatically stop teaching the human since the human's internal model is sufficiently correct (supporting \textbf{H3}). 
The one exception is in the \textbf{Robot Arm Teleoperation} environment with the \textbf{threshold} human. Since this human doesn't learn when the gradient is too small, the robot must continue to exert effort to maximize its reward. 
\vspace{-0.3cm}

\subsection{Implicitly Influencing Human Objectives}
\label{subsec:reward_manipulation}

We now turn to scenarios where the human has an accurate understanding of the robot's dynamics, but their objective (i.e., their reward function $r_\mathrm{H}$) can be changed by the robot.
Specifically, we study how assistive robots can implicitly influence human motion preferences and desired goals. 
Importantly, in this setting influencing or teaching the human is not explicitly in the robot's objective: the robot simply wants to perform the desired task with minimal assistance. Thus, getting the human to want to reach a goal or change their preferences should be an emergent behavior of robots planning with the dynamics of human learning. 


\subsubsection{Dynamics of human learning.}
\label{subsubsec:sim_humans_reward}
We simulate\footnote{While we simulate the human as changing their reward, but the human's reward could be viewed as static while their subgoals change
Nonetheless, it will be common for a robot to not fully represent this hierarchy. 
} the \textbf{gradient} human learner from \ref{subsubsec:sim_humans_phys} and introduce a new human, the \textbf{Bayesian} learner\footnote{\textbf{Bayesian} humans act under their belief: $\mathbb{P}(\uH \mid \jointstate) = \sum_{\param} b(\param) \mathbb{P}(\uH \mid \jointstate, \param)$.} \edit{which is inspired by probabilistic models of cognition \cite{griffiths2010probabilistic,annurev-devpsych-121318-084833}}.
This human's learning produces a full posterior, $b^{t+1}(\param)$, over the model parameters given a state-action observation, and the dynamics of learning are:
$\paramdyn^{\textrm{Bayes}} \propto P(u^t \mid \jointstate^t, \param)b^t(\param)$. 

\subsubsection{Human internal model.}
Since the human's objectives are influenceable, we model $\param$ as a reward parameter encoding the motion preferences $\param := (Q, R)$ or a desired goal state $\param \in \Theta$.  

\subsubsection{Simulated environments.}
We assume the human knows the physical robot dynamics (the bias-free \textbf{RobotArm} dynamics from \ref{subsubsec:phys_dyns_sim_envs}), but can have their reward influenced by new observations. 

\smallskip
\noindent \textbf{(1) Goal Influence.}
The human wants to teleoperate the robot to put an object in one of the three trays (upper left \figref{fig:goal_and_pref_results}). However,the human doesn't notice that only one of the trays is empty enough. 
Unlike the human, the robot's sensors detect that only one of the trays is empty.
We investigate if the robot can influence the human to change their preferences about which tray (i.e., goal location) to place their object in. 

\smallskip
\noindent \textbf{(2) Preference Influence.}
The human wants to teleoperate the robot to pick up a cup on the table. \edit{Their} initial preference is to move the robot's end-effector in a straight line from start to the cup (lower left \figref{fig:goal_and_pref_results}). However, the robot knows that grasps tend to fail with this kind of motion. 
Instead, the robot knows that first moving \edit{directly above the can} and then straight down to \edit{grasp} has a \edit{higher} chance of success. 
We investigate if the robot can influence the human to change their preferences about how to \edit{reach the cup}. 


\subsubsection{Human objective.}
In all simulations the human has a quadratic cost function (from \eqref{eq:human_reward_quadratic}). In \textbf{Goal Influence} the simulated human receives reward for moving the robot end-effector to their desired tray, and in \textbf{Preference Influence} the human receives reward according to their current preference matricies, $(Q,R)$.

\subsubsection{Robot objective.}
We implement an \textit{assistive robot} that wants to help the human perform the task while minimally intervening. However, we assume that the robot knows best: the robot knows which goal or reward weights lead to success. Let $\theta^*$ capture this aspect of the robot's reward.
In the \textbf{Goal} setting the robot's reward
$r_\mathrm{R}(s, \uR, \uH; \theta^*) = -(\jointstate-\theta^*)^\top Q (\jointstate-\theta^*) - u^\top R u - ||u-\uH||^2_2$ and in \textbf{Preference} the robot's reward parameter is $\theta^*=(Q^*, R^*)$, yielding  $r_\mathrm{R}(s, \uR, \uH; \theta^*) = -\jointstate^\top Q^* \jointstate - u^\top R^* u - ||u-\uH||^2_2$ where $u$ is the combined human and robot action from Equation~\eqref{eq:shared_control}.

\subsubsection{Baselines.}
We implement our method where the robot assists the human and plans with the inferred dynamics of human learning (\textbf{Learning Assist}). We compare to a robot assisting with the ground-truth dynamics of human learning (\textbf{Oracle}), robot assistance that is unaware that humans learn (\textbf{Static Assist}), and a robot that randomly perturbs the human actions (\textbf{Random}).

\subsubsection{Hypotheses.}
\textbf{H\showHnum:} \textit{\textbf{Learning Assist} aligns the human's mental model faster}.
\textbf{H\showHnum:} \textit{Assistance that accounts for human learning enables the human-robot team to achieve higher reward under the true $\theta^*$.}

\subsubsection{Results.}
\begin{figure}[t!]
    \centering
    \includegraphics[width=0.48\textwidth]{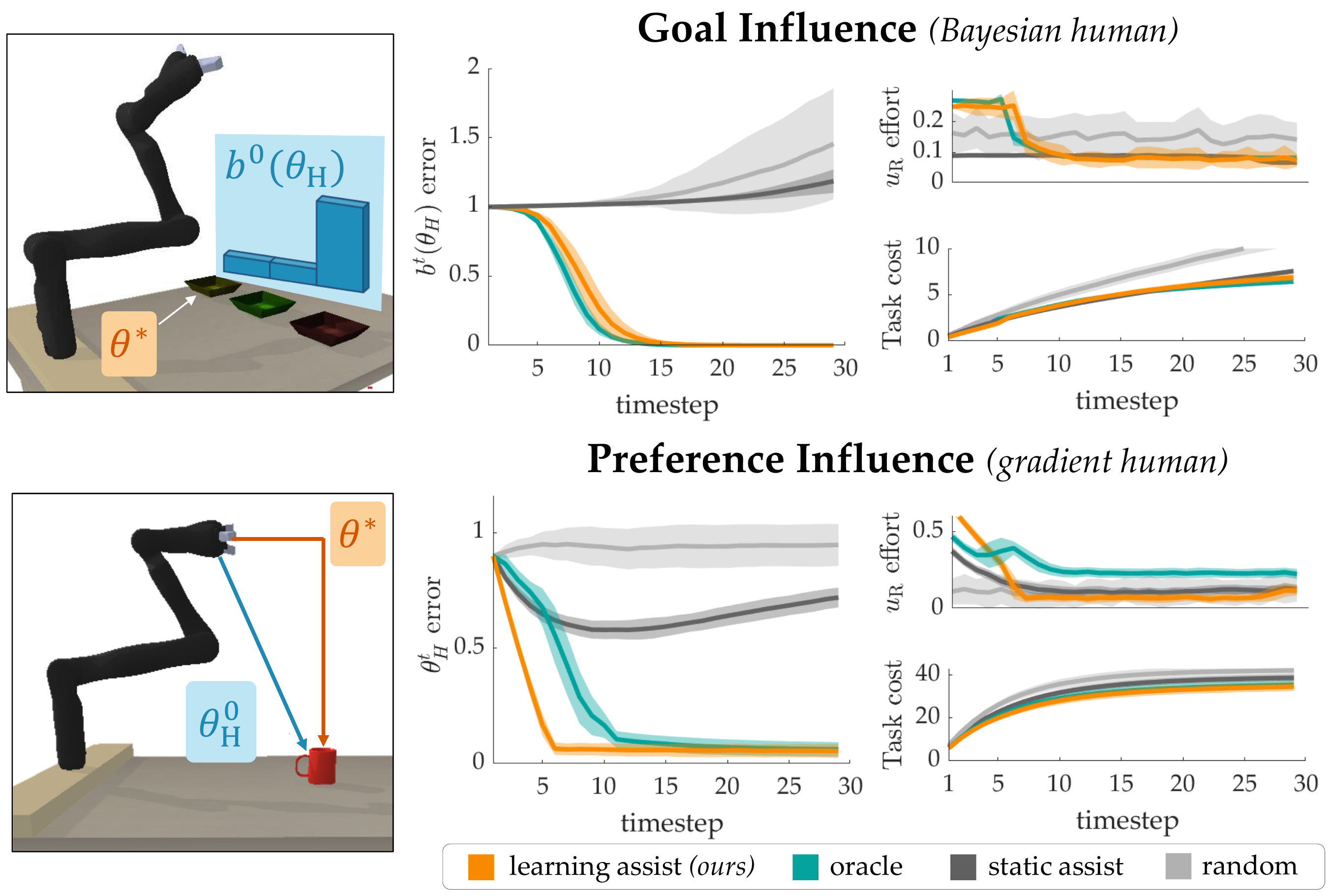}
    \vspace{-2em}
    \caption{(left) Environments for influencing human objectives. (right) Internal model error, robot effort, and task cost.}
    \label{fig:goal_and_pref_results}
    \vspace{-1.8em}
\end{figure}

\figref{fig:goal_and_pref_results} shows the human's internal model error, robot effort, and task cost (i.e., just the task-component of $r_\mathrm{R}$, negated) for both environments. Because the \textbf{Learning Assist} robot knows that the human's internal model can be changed, it \textit{automatically} exerts higher effort early on to align the human's internal model with it's own, resulting in less long-term assistance and lower task cost (supporting \textbf{H4} and \textbf{H5}). 
In contrast, the \textbf{Static Assist} robot
is not aware that the human can change their mind, and \edit{thus} does not \edit{exert} enough effort to influence the human's internal model. 
After repeatedly incurring task cost because the two agents are at odds with each other,
the \textbf{Static Assist} robot ``gives up'' and starts executing the human's control directly: in other words, $u = \uH$. 






\section{User Study: Teaching to Teleoperate}
\label{sec:user_study}

So far we conducted experiments with simulated human behavior, allowing us to analyze the quality of our inferred human learning dynamics model, and the robot's ability to influence simulated humans.
Here we investigate if we can infer the dynamics of \textit{real} human learning, and enable robots to influence real users. 

We focus on scenarios where the robot's physical dynamics are different from what the human is used to; for example, perhaps the human was used to teleoperating a robotic wheelchair, but is now teleoperating a robotic arm.
As they interact with the robotic arm, they will naturally learn about the new robot dynamics. 
In our IRB-approved user study, we investigate if a robot can actively teach a human the physical dynamics and improve their teleoperation performance faster than if the human does the task on their own. 
In other words, we aim to understand if a robot can \textit{align} the human's internal model with the robot's.

\medskip 
\noindent \textbf{Experimental Setup.}
We designed a teleoperation task where the human controls a 7DOF
Jaco robot arm through a webcam-based gesture interface (\figref{fig:front_fig}).
The participant uses their index finger to indicate how the end-effector should move 
parallel to the tabletop. 
The task is to move the end-effector to reach four goals on the table in a counter-clockwise pattern, tracing out a diamond pattern.
All participants experience a familiarzation task where they perform the task unassisted, with the default robot dynamics in order to understand the gesture interface.
In software, we then simulate two ``new'' robots, each with different physical properties.


\medskip 
\noindent \textbf{Independent Variables.} 
We manipulated the \textit{robot strategy} with two levels: \textit{no-teaching} and \textit{active-teaching}. The robot either let the human do the task on their own, or it modified the human's input to teach them about the physical robot dynamics via Equation \eqref{eq:robot_planning_problem}. 
We also manipulate the \textit{robot physical dynamics} with two levels: end-effector dynamics \textit{bias in x-direction} and \textit{bias in y-direction}.

\medskip 
\noindent \textbf{Dependent Measures.} 
A challenge in evaluating our experiment is that we do not have access to the human's ground-truth internal model.
As a proxy, we measure \textit{human action optimality distance}: $||\hat{u}_H - u^*||^2_2$. \edit{Intuitively, the better the human understands the robot, the more optimally they should be able to control it to reach the goals. Since we cannot directly measure a human’s internal understanding, we instead look at their actions to measure their deviation from the optimal action under the robot’s true physics.}
We also measured subjective measures via a Likert scale survey.

\medskip 
\noindent \textbf{Hypotheses.} \textbf{H\showHnum:} \textit{Participants in the active teaching condition become optimal teleoperators faster than passively learning on their own.} \textbf{H\showHnum:} \textit{Participants feel they learned to teleoperate faster and understood the robot dynamics better in the active teaching condition.}

\medskip 
\noindent \textbf{Participants.}
We recruited two groups of participants from the campus community: the first for providing data for inferring the dynamics of human learning (12 participants; 2 female, 10 male, age 18-34, all with technical backgrounds), and the second for the user study (10 participants; 1 female, 8 male, 1 non-binary, age 18-34, all with technical backgrounds).
For inferring the human learning dynamics, all participants learned to teleoperate the robot unassisted and we counterbalanced the \textit{robot physical dynamics}.

\medskip 
\noindent \textbf{Procedure.} 
A within-subjects design is challenging, since humans who experience one condition will learn about the \edit{robots and} then carry over that experience to the next condition. 
To study the effect of this confound, each participant experienced a combination of \textit{robot strategy} and \textit{physical dynamics} conditions, but in a random order. For example, one group of participants would interact with the \textit{(active-teaching, bias-x)} condition and then \textit{(no-teaching, bias-y)} condition. 
Thus, each participant experiences both robot strategies and biases. 
We counterbalance the order in which the participants experience the combination. 
All participants experienced a familiarization round at the start and between each experimental condition, to ``reset'' their mental model of the robot. \edit{Each participant gave 3 demonstrations per condition, each lasting $\sim$1 minute.}

\begin{figure}[t!]
    \centering
    \includegraphics[width=0.48\textwidth]{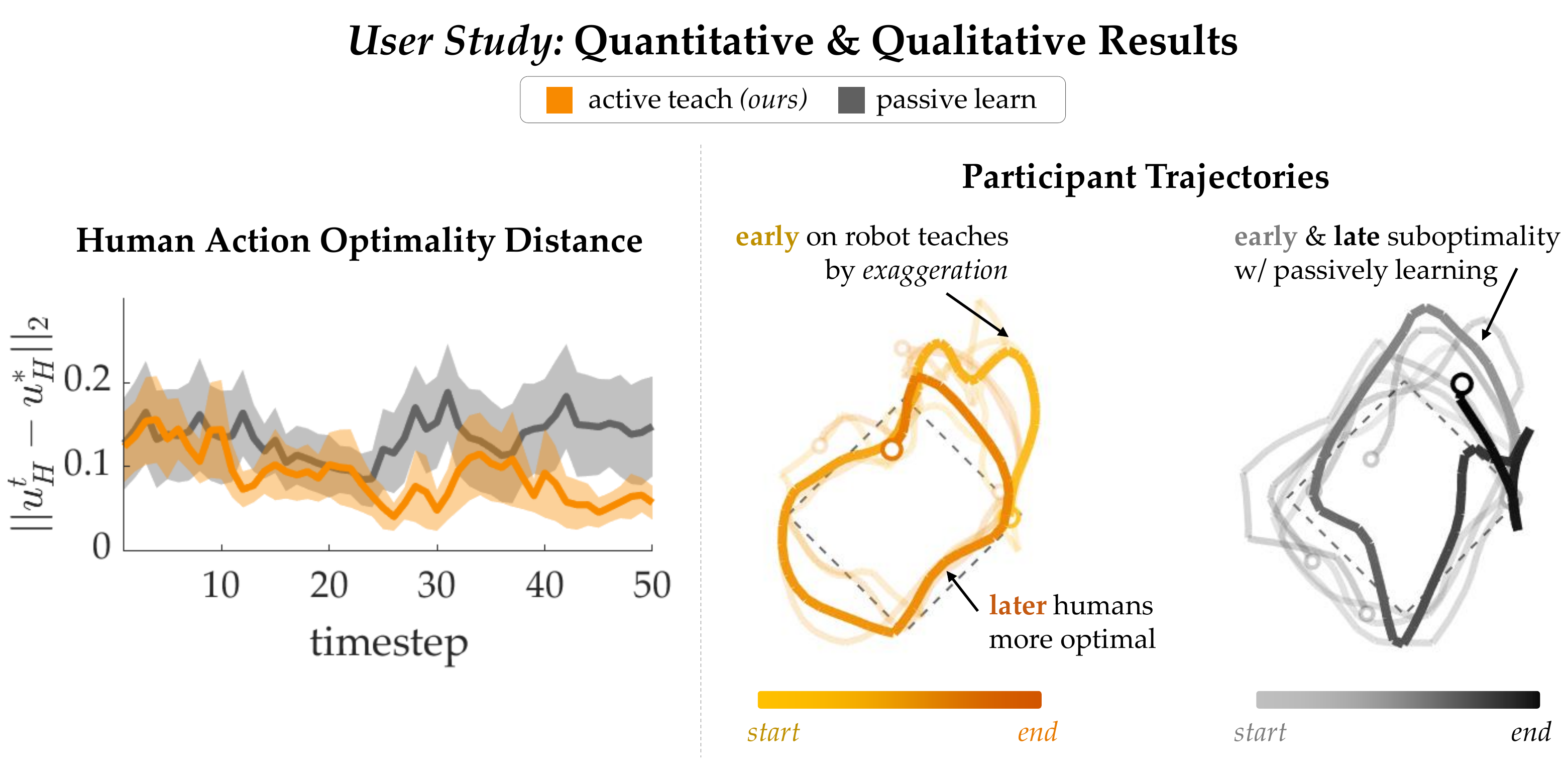}
    \vspace{-0.8cm}
    \caption{(left) Avg. human action optimality distance and 95\% confidence interval.
    (right) Dashed line is desired path. Participant trajectories reveal that an active teaching robot initially \textit{exaggerates} the dynamics bias to teach the human.}
    \label{fig:user_study_results}
    \vspace{-1.8em}
\end{figure}

\medskip
\noindent \textbf{Quantitative Results.}
Figure \ref{fig:user_study_results} shows how \textit{human action optimality distance} varies over time with each robot strategy. We conducted an ANOVA with \textit{robot strategy} and stage (first or second half of interaction) as factors and \textit{robot physical dynamics} as random effect. 
We found a significant main effect of the robot strategy ($F(1, 19) = 12.943, p = 0.001$) and a marginal interaction effect between the robot strategy and the interaction stage ($p = 0.098$), so we did not run a post-hoc analysis. However, we hypothesize that this marginal interaction effect comes from the fact that early-stage changes in robot behavior (induced by either robot strategy) influences the human's later-stage action optimality. Ultimately, the quantitative results indicate a significant improvement in the human's action optimality when the robot actively teaches them compared to when the human passively learns (supporting \textbf{H6}).

\medskip
\noindent \textbf{Qualitative \& Subjective Results.}
On the right of \figref{fig:user_study_results} we visualize the executed trajectories from all participants in the active-teaching (orange) and no-teaching (grey) conditions. 
The highlighted trajectories are two representative examples, the color gradient indicates time along the trajectory, and the dashed line is the desired path.  
When participants passively learn on their own, their trajectories are consistently suboptimal, weaving around the optimal path. In contrast, in the active teaching condition, the initial portion of the trajectory exhibits the robots teaching behavior: the robot intentionally \textit{exaggerates} the dynamics bias to change the human's internal model faster. After this initial exaggerated deviation, the human trajectory is closer to optimal compared to the passive learning trajectory at comparable timesteps  \edit{(see Appendix~ \ref{app:user_study_alignment} for 
a detailed visualization of human and robot actions).}

We also ran an ANOVA on the Likert survey questions. Survey questions investigated perceived performance improvements (e.g., ``By the end of the interaction, it was easy to control the robot to do the task.'') and \edit{robot understanding} (e.g., ``By the end of the interaction, I understood the robot's physical properties.''). Across all questions, we did not find a significant effect of the robot strategy (rejecting \textbf{H7}).
What we found surprising was that even though participants were \textit{quantitatively} performing better in the teaching condition, they did not \textit{perceive} an improvement in performance ($p = 0.689$) nor in their understanding of the robot physics ($p = 0.299$). 
We hypothesize that this could be because participants 
only interacted with each robot strategy for one minute, making the differences hard to notice. In the future, investigating longer-term interactions with the robot would shed light on the disconnect.



\section{Conclusion}
In this work \edit{we took a step towards enabling} robots to understand the influence that they have over human internal models.
We do this by \edit{modeling} human learning as a nonlinear dynamical system that evolves as a function of new observations that the robot can influence. 
We propose a tractable method for inferring \edit{approximate} human learning dynamics from demonstrations that naturally exhibit human learning, and propose how robots can influence human learning by embedding the \edit{approximate dynamics} into robot planning. 
\edit{Our experimental results indicate that robot influence is possible and can help humans learn better internal models.}

\smallskip 
\noindent \textbf{Limitations \& Future Work.}
\edit{A strength and limitation of our approach is representing the dynamics of human learning via a transformer. As a general function approximator, it poses no assumptions on the structure of the human’s learning dynamics; in fact, we are excited that our results indicate that it is possible to infer a useful model of human learning from real data, without prior assumptions. 
However, since neural networks require abundant human data, they are not appropriate for low-data settings and may fail when encountering humans that are out of distribution. 
A further limitation is that if the person is not noisily-optimal as in \eqref{eq:human_policy} and has a specific bias (e.g., myopia), then the transformer will learn parameters that compensate for this; in turn, this could lead the robot to influence the human in unintended ways. 
In the future we are excited to combine the strengths of data-driven models and cognitive science models of human learning.
While our user study relies on an ``average'' dynamics model of human learning trained from all participants' data, humans may exhibit unique ways of learning. Inferring personalized learning dynamics is an exciting future direction, and pre-trained models of humans could serve as a useful starting point for adapting to new humans.
Finally,} while the LQ approximation enables tractable inference, extensions into non-LQ settings \edit{will unlock more settings (e.g., autonomous cars).} 
\bibliographystyle{ACM-Reference-Format}
\bibliography{sample-base}


\begin{thebibliography}{55}


\ifx \showCODEN    \undefined \def \showCODEN     #1{\unskip}     \fi
\ifx \showDOI      \undefined \def \showDOI       #1{#1}\fi
\ifx \showISBNx    \undefined \def \showISBNx     #1{\unskip}     \fi
\ifx \showISBNxiii \undefined \def \showISBNxiii  #1{\unskip}     \fi
\ifx \showISSN     \undefined \def \showISSN      #1{\unskip}     \fi
\ifx \showLCCN     \undefined \def \showLCCN      #1{\unskip}     \fi
\ifx \shownote     \undefined \def \shownote      #1{#1}          \fi
\ifx \showarticletitle \undefined \def \showarticletitle #1{#1}   \fi
\ifx \showURL      \undefined \def \showURL       {\relax}        \fi
\providecommand\bibfield[2]{#2}
\providecommand\bibinfo[2]{#2}
\providecommand\natexlab[1]{#1}
\providecommand\showeprint[2][]{arXiv:#2}

\bibitem[Admoni and Scassellati(2017)]%
        {admoni2017social}
\bibfield{author}{\bibinfo{person}{Henny Admoni} {and} \bibinfo{person}{Brian
  Scassellati}.} \bibinfo{year}{2017}\natexlab{}.
\newblock \showarticletitle{Social eye gaze in human-robot interaction: a
  review}.
\newblock \bibinfo{journal}{\emph{Journal of Human-Robot Interaction}}
  \bibinfo{volume}{6}, \bibinfo{number}{1} (\bibinfo{year}{2017}),
  \bibinfo{pages}{25--63}.
\newblock


\bibitem[Baker et~al\mbox{.}(2009)]%
        {baker2009action}
\bibfield{author}{\bibinfo{person}{Chris~L Baker}, \bibinfo{person}{Rebecca
  Saxe}, {and} \bibinfo{person}{Joshua~B Tenenbaum}.}
  \bibinfo{year}{2009}\natexlab{}.
\newblock \showarticletitle{Action understanding as inverse planning}.
\newblock \bibinfo{journal}{\emph{Cognition}} \bibinfo{volume}{113},
  \bibinfo{number}{3} (\bibinfo{year}{2009}), \bibinfo{pages}{329--349}.
\newblock


\bibitem[Bertsekas et~al\mbox{.}(2011)]%
        {bertsekas2011dynamic}
\bibfield{author}{\bibinfo{person}{Dimitri~P Bertsekas} {et~al\mbox{.}}}
  \bibinfo{year}{2011}\natexlab{}.
\newblock \showarticletitle{Dynamic programming and optimal control 3rd
  edition, volume ii}.
\newblock \bibinfo{journal}{\emph{Belmont, MA: Athena Scientific}}
  (\bibinfo{year}{2011}).
\newblock


\bibitem[Cakmak and Lopes(2012)]%
        {cakmak2012algorithmic}
\bibfield{author}{\bibinfo{person}{Maya Cakmak} {and} \bibinfo{person}{Manuel
  Lopes}.} \bibinfo{year}{2012}\natexlab{}.
\newblock \showarticletitle{Algorithmic and human teaching of sequential
  decision tasks}. In \bibinfo{booktitle}{\emph{Conference on Artificial
  Intelligence}}.
\newblock


\bibitem[Camacho and Alba(2013)]%
        {camacho2013model}
\bibfield{author}{\bibinfo{person}{Eduardo~F Camacho} {and}
  \bibinfo{person}{Carlos~Bordons Alba}.} \bibinfo{year}{2013}\natexlab{}.
\newblock \bibinfo{booktitle}{\emph{Model predictive control}}.
\newblock \bibinfo{publisher}{Springer science \& business media}.
\newblock


\bibitem[Chan et~al\mbox{.}(2019)]%
        {chan2019assistive}
\bibfield{author}{\bibinfo{person}{Lawrence Chan}, \bibinfo{person}{Dylan
  Hadfield-Menell}, \bibinfo{person}{Siddhartha Srinivasa}, {and}
  \bibinfo{person}{Anca Dragan}.} \bibinfo{year}{2019}\natexlab{}.
\newblock \showarticletitle{The assistive multi-armed bandit}. In
  \bibinfo{booktitle}{\emph{2019 14th ACM/IEEE International Conference on
  Human-Robot Interaction (HRI)}}. IEEE, \bibinfo{pages}{354--363}.
\newblock


\bibitem[Chen et~al\mbox{.}(2020)]%
        {chen2020trust}
\bibfield{author}{\bibinfo{person}{Min Chen}, \bibinfo{person}{Stefanos
  Nikolaidis}, \bibinfo{person}{Harold Soh}, \bibinfo{person}{David Hsu}, {and}
  \bibinfo{person}{Siddhartha Srinivasa}.} \bibinfo{year}{2020}\natexlab{}.
\newblock \showarticletitle{Trust-aware decision making for human-robot
  collaboration: Model learning and planning}.
\newblock \bibinfo{journal}{\emph{ACM Transactions on Human-Robot Interaction
  (THRI)}} (\bibinfo{year}{2020}).
\newblock


\bibitem[Dragan et~al\mbox{.}(2013)]%
        {dragan2013legibility}
\bibfield{author}{\bibinfo{person}{Anca~D Dragan}, \bibinfo{person}{Kenton~CT
  Lee}, {and} \bibinfo{person}{Siddhartha~S Srinivasa}.}
  \bibinfo{year}{2013}\natexlab{}.
\newblock \showarticletitle{Legibility and predictability of robot motion}. In
  \bibinfo{booktitle}{\emph{2013 8th ACM/IEEE International Conference on
  Human-Robot Interaction (HRI)}}. IEEE, \bibinfo{pages}{301--308}.
\newblock


\bibitem[Dragan and Srinivasa(2013)]%
        {dragan2013policy}
\bibfield{author}{\bibinfo{person}{Anca~D Dragan} {and}
  \bibinfo{person}{Siddhartha~S Srinivasa}.} \bibinfo{year}{2013}\natexlab{}.
\newblock \showarticletitle{A policy-blending formalism for shared control}.
\newblock \bibinfo{journal}{\emph{The International Journal of Robotics
  Research}} \bibinfo{volume}{32}, \bibinfo{number}{7} (\bibinfo{year}{2013}),
  \bibinfo{pages}{790--805}.
\newblock


\bibitem[East et~al\mbox{.}(2020)]%
        {east2020infinite}
\bibfield{author}{\bibinfo{person}{Sebastian East}, \bibinfo{person}{Marco
  Gallieri}, \bibinfo{person}{Jonathan Masci}, \bibinfo{person}{Jan
  Koutn{\'\i}k}, {and} \bibinfo{person}{Mark Cannon}.}
  \bibinfo{year}{2020}\natexlab{}.
\newblock \showarticletitle{Infinite-horizon differentiable model predictive
  control}.
\newblock \bibinfo{journal}{\emph{International Conference on Learning
  Representations}} (\bibinfo{year}{2020}).
\newblock


\bibitem[Face(2022)]%
        {hugging2022transformer}
\bibfield{author}{\bibinfo{person}{Hugging Face}.}
  \bibinfo{year}{2022}\natexlab{}.
\newblock \bibinfo{booktitle}{\emph{Transformers}}.
\newblock
\urldef\tempurl%
\url{https://huggingface.co/docs/transformers/index}
\showURL{%
\tempurl}


\bibitem[Goodman et~al\mbox{.}(2016)]%
        {goodman2016probabilistic}
\bibfield{author}{\bibinfo{person}{Noah~D Goodman}, \bibinfo{person}{Joshua~B.
  Tenenbaum}, {and} \bibinfo{person}{The~ProbMods Contributors}.}
  \bibinfo{year}{2016}\natexlab{}.
\newblock \bibinfo{title}{{Probabilistic Models of Cognition}}.
\newblock \bibinfo{howpublished}{\url{http://probmods.org/v2}}.
\newblock
\newblock
\shownote{Accessed: 2022-12-9}.


\bibitem[Griffiths et~al\mbox{.}(2010)]%
        {griffiths2010probabilistic}
\bibfield{author}{\bibinfo{person}{Thomas~L Griffiths}, \bibinfo{person}{Nick
  Chater}, \bibinfo{person}{Charles Kemp}, \bibinfo{person}{Amy Perfors}, {and}
  \bibinfo{person}{Joshua~B Tenenbaum}.} \bibinfo{year}{2010}\natexlab{}.
\newblock \showarticletitle{Probabilistic models of cognition: Exploring
  representations and inductive biases}.
\newblock \bibinfo{journal}{\emph{Trends in cognitive sciences}}
  \bibinfo{volume}{14}, \bibinfo{number}{8} (\bibinfo{year}{2010}),
  \bibinfo{pages}{357--364}.
\newblock


\bibitem[Habibian and Losey(2022)]%
        {habibian2022encouraging}
\bibfield{author}{\bibinfo{person}{Soheil Habibian} {and}
  \bibinfo{person}{Dylan~P. Losey}.} \bibinfo{year}{2022}\natexlab{}.
\newblock \showarticletitle{Encouraging Human Interaction with Robot Teams:
  Legible and Fair Subtask Allocations}.
\newblock \bibinfo{journal}{\emph{IEEE Robotics and Automation Letters}}
  \bibinfo{volume}{8} (\bibinfo{year}{2022}), \bibinfo{pages}{188}.
\newblock


\bibitem[Hu and Fisac(2022)]%
        {hu2022active}
\bibfield{author}{\bibinfo{person}{Haimin Hu} {and} \bibinfo{person}{Jaime~F
  Fisac}.} \bibinfo{year}{2022}\natexlab{}.
\newblock \showarticletitle{Active Uncertainty Reduction for Human-Robot
  Interaction: An Implicit Dual Control Approach}.
\newblock \bibinfo{journal}{\emph{WAFR}} (\bibinfo{year}{2022}).
\newblock


\bibitem[Huang et~al\mbox{.}(2019)]%
        {huang2019enabling}
\bibfield{author}{\bibinfo{person}{Sandy~H Huang}, \bibinfo{person}{David
  Held}, \bibinfo{person}{Pieter Abbeel}, {and} \bibinfo{person}{Anca~D
  Dragan}.} \bibinfo{year}{2019}\natexlab{}.
\newblock \showarticletitle{Enabling robots to communicate their objectives}.
\newblock \bibinfo{journal}{\emph{Autonomous Robots}} \bibinfo{volume}{43},
  \bibinfo{number}{2} (\bibinfo{year}{2019}), \bibinfo{pages}{309--326}.
\newblock


\bibitem[Jain and Argall(2019)]%
        {jain2019probabilistic}
\bibfield{author}{\bibinfo{person}{Siddarth Jain} {and} \bibinfo{person}{Brenna
  Argall}.} \bibinfo{year}{2019}\natexlab{}.
\newblock \showarticletitle{Probabilistic human intent recognition for shared
  autonomy in assistive robotics}.
\newblock \bibinfo{journal}{\emph{ACM Transactions on Human-Robot Interaction
  (THRI)}} \bibinfo{volume}{9}, \bibinfo{number}{1} (\bibinfo{year}{2019}),
  \bibinfo{pages}{1--23}.
\newblock


\bibitem[Janner et~al\mbox{.}(2021)]%
        {janner2021offline}
\bibfield{author}{\bibinfo{person}{Michael Janner}, \bibinfo{person}{Qiyang
  Li}, {and} \bibinfo{person}{Sergey Levine}.} \bibinfo{year}{2021}\natexlab{}.
\newblock \showarticletitle{Offline reinforcement learning as one big sequence
  modeling problem}.
\newblock \bibinfo{journal}{\emph{Advances in neural information processing
  systems}}  \bibinfo{volume}{34} (\bibinfo{year}{2021}),
  \bibinfo{pages}{1273--1286}.
\newblock


\bibitem[Jara-Ettinger(2019)]%
        {jara2019theory}
\bibfield{author}{\bibinfo{person}{Julian Jara-Ettinger}.}
  \bibinfo{year}{2019}\natexlab{}.
\newblock \showarticletitle{Theory of mind as inverse reinforcement learning}.
\newblock \bibinfo{journal}{\emph{Current Opinion in Behavioral Sciences}}
  \bibinfo{volume}{29} (\bibinfo{year}{2019}), \bibinfo{pages}{105--110}.
\newblock


\bibitem[Kalman et~al\mbox{.}(1960)]%
        {kalman1960contributions}
\bibfield{author}{\bibinfo{person}{Rudolf~Emil Kalman} {et~al\mbox{.}}}
  \bibinfo{year}{1960}\natexlab{}.
\newblock \showarticletitle{Contributions to the theory of optimal control}.
\newblock \bibinfo{journal}{\emph{Bol. soc. mat. mexicana}}
  \bibinfo{volume}{5}, \bibinfo{number}{2} (\bibinfo{year}{1960}),
  \bibinfo{pages}{102--119}.
\newblock


\bibitem[Khalil and Dombre(2002)]%
        {khalil2002modeling}
\bibfield{author}{\bibinfo{person}{Wisama Khalil} {and}
  \bibinfo{person}{Etienne Dombre}.} \bibinfo{year}{2002}\natexlab{}.
\newblock \bibinfo{booktitle}{\emph{Modeling identification and control of
  robots}}.
\newblock \bibinfo{publisher}{CRC Press}.
\newblock


\bibitem[Kitani et~al\mbox{.}(2012)]%
        {kitani2012activity}
\bibfield{author}{\bibinfo{person}{Kris~M Kitani}, \bibinfo{person}{Brian~D
  Ziebart}, \bibinfo{person}{James~Andrew Bagnell}, {and}
  \bibinfo{person}{Martial Hebert}.} \bibinfo{year}{2012}\natexlab{}.
\newblock \showarticletitle{Activity forecasting}. In
  \bibinfo{booktitle}{\emph{European conference on computer vision}}. Springer,
  \bibinfo{pages}{201--214}.
\newblock


\bibitem[Laine et~al\mbox{.}(2021)]%
        {laine2021multi}
\bibfield{author}{\bibinfo{person}{Forrest Laine}, \bibinfo{person}{David
  Fridovich-Keil}, \bibinfo{person}{Chih-Yuan Chiu}, {and}
  \bibinfo{person}{Claire Tomlin}.} \bibinfo{year}{2021}\natexlab{}.
\newblock \showarticletitle{Multi-Hypothesis Interactions in Game-Theoretic
  Motion Planning}. In \bibinfo{booktitle}{\emph{2021 IEEE International
  Conference on Robotics and Automation (ICRA)}}. IEEE,
  \bibinfo{pages}{8016--8023}.
\newblock


\bibitem[Levine and Koltun(2012)]%
        {levine2012continuous}
\bibfield{author}{\bibinfo{person}{Sergey Levine} {and}
  \bibinfo{person}{Vladlen Koltun}.} \bibinfo{year}{2012}\natexlab{}.
\newblock \showarticletitle{Continuous inverse optimal control with locally
  optimal examples}.
\newblock \bibinfo{journal}{\emph{arXiv preprint arXiv:1206.4617}}
  (\bibinfo{year}{2012}).
\newblock


\bibitem[Li et~al\mbox{.}(2021)]%
        {li2021roial}
\bibfield{author}{\bibinfo{person}{Kejun Li}, \bibinfo{person}{Maegan Tucker},
  \bibinfo{person}{Erdem B{\i}y{\i}k}, \bibinfo{person}{Ellen Novoseller},
  \bibinfo{person}{Joel~W Burdick}, \bibinfo{person}{Yanan Sui},
  \bibinfo{person}{Dorsa Sadigh}, \bibinfo{person}{Yisong Yue}, {and}
  \bibinfo{person}{Aaron~D Ames}.} \bibinfo{year}{2021}\natexlab{}.
\newblock \showarticletitle{Roial: Region of interest active learning for
  characterizing exoskeleton gait preference landscapes}. In
  \bibinfo{booktitle}{\emph{2021 IEEE International Conference on Robotics and
  Automation (ICRA)}}. IEEE, \bibinfo{pages}{3212--3218}.
\newblock


\bibitem[Losey et~al\mbox{.}(2022)]%
        {losey2022learning}
\bibfield{author}{\bibinfo{person}{Dylan~P Losey}, \bibinfo{person}{Hong~Jun
  Jeon}, \bibinfo{person}{Mengxi Li}, \bibinfo{person}{Krishnan Srinivasan},
  \bibinfo{person}{Ajay Mandlekar}, \bibinfo{person}{Animesh Garg},
  \bibinfo{person}{Jeannette Bohg}, {and} \bibinfo{person}{Dorsa Sadigh}.}
  \bibinfo{year}{2022}\natexlab{}.
\newblock \showarticletitle{Learning latent actions to control assistive
  robots}.
\newblock \bibinfo{journal}{\emph{Autonomous robots}} \bibinfo{volume}{46},
  \bibinfo{number}{1} (\bibinfo{year}{2022}), \bibinfo{pages}{115--147}.
\newblock


\bibitem[Marr(1982)]%
        {marr2010vision}
\bibfield{author}{\bibinfo{person}{David Marr}.}
  \bibinfo{year}{1982}\natexlab{}.
\newblock \bibinfo{booktitle}{\emph{Vision: A computational investigation into
  the human representation and processing of visual information}}.
\newblock \bibinfo{publisher}{W.H. Freeman}.
\newblock


\bibitem[Mulder(1999)]%
        {mulder1999cybernetics}
\bibfield{author}{\bibinfo{person}{Max Mulder}.}
  \bibinfo{year}{1999}\natexlab{}.
\newblock \showarticletitle{Cybernetics of tunnel-in-the-sky displays}.
\newblock  (\bibinfo{year}{1999}).
\newblock


\bibitem[Newman et~al\mbox{.}(2022)]%
        {newman2022harmonic}
\bibfield{author}{\bibinfo{person}{Benjamin~A Newman},
  \bibinfo{person}{Reuben~M Aronson}, \bibinfo{person}{Siddhartha~S Srinivasa},
  \bibinfo{person}{Kris Kitani}, {and} \bibinfo{person}{Henny Admoni}.}
  \bibinfo{year}{2022}\natexlab{}.
\newblock \showarticletitle{HARMONIC: A multimodal dataset of assistive
  human--robot collaboration}.
\newblock \bibinfo{journal}{\emph{The International Journal of Robotics
  Research}} (\bibinfo{year}{2022}).
\newblock


\bibitem[Newman et~al\mbox{.}(2020)]%
        {newman2020examining}
\bibfield{author}{\bibinfo{person}{Benjamin~A Newman}, \bibinfo{person}{Abhijat
  Biswas}, \bibinfo{person}{Sarthak Ahuja}, \bibinfo{person}{Siddharth
  Girdhar}, \bibinfo{person}{Kris~K Kitani}, {and} \bibinfo{person}{Henny
  Admoni}.} \bibinfo{year}{2020}\natexlab{}.
\newblock \showarticletitle{Examining the effects of anticipatory robot
  assistance on human decision making}. In
  \bibinfo{booktitle}{\emph{International Conference on Social Robotics}}.
  Springer, \bibinfo{pages}{590--603}.
\newblock


\bibitem[Ng et~al\mbox{.}(2000)]%
        {ng2000algorithms}
\bibfield{author}{\bibinfo{person}{Andrew~Y Ng}, \bibinfo{person}{Stuart~J
  Russell}, {et~al\mbox{.}}} \bibinfo{year}{2000}\natexlab{}.
\newblock \showarticletitle{Algorithms for inverse reinforcement learning.}. In
  \bibinfo{booktitle}{\emph{Icml}}, Vol.~\bibinfo{volume}{1}.
  \bibinfo{pages}{2}.
\newblock


\bibitem[Nikolaidis et~al\mbox{.}(2017)]%
        {nikolaidis2017game}
\bibfield{author}{\bibinfo{person}{Stefanos Nikolaidis},
  \bibinfo{person}{Swaprava Nath}, \bibinfo{person}{Ariel~D Procaccia}, {and}
  \bibinfo{person}{Siddhartha Srinivasa}.} \bibinfo{year}{2017}\natexlab{}.
\newblock \showarticletitle{Game-theoretic modeling of human adaptation in
  human-robot collaboration}. In \bibinfo{booktitle}{\emph{Proceedings of the
  2017 ACM/IEEE international conference on human-robot interaction}}.
  \bibinfo{pages}{323--331}.
\newblock


\bibitem[Parekh et~al\mbox{.}(2022)]%
        {parekh2022rili}
\bibfield{author}{\bibinfo{person}{Sagar Parekh}, \bibinfo{person}{Soheil
  Habibian}, {and} \bibinfo{person}{Dylan~P Losey}.}
  \bibinfo{year}{2022}\natexlab{}.
\newblock \showarticletitle{RILI: Robustly Influencing Latent Intent}.
\newblock \bibinfo{journal}{\emph{arXiv preprint arXiv:2203.12705}}
  (\bibinfo{year}{2022}).
\newblock


\bibitem[Peters et~al\mbox{.}(2021)]%
        {peters2021inferring}
\bibfield{author}{\bibinfo{person}{Lasse Peters}, \bibinfo{person}{David
  Fridovich-Keil}, \bibinfo{person}{Vicen{\c{c}} Rubies-Royo},
  \bibinfo{person}{Claire~J Tomlin}, {and} \bibinfo{person}{Cyrill Stachniss}.}
  \bibinfo{year}{2021}\natexlab{}.
\newblock \showarticletitle{Inferring objectives in continuous dynamic games
  from noise-corrupted partial state observations}.
\newblock \bibinfo{journal}{\emph{arXiv preprint arXiv:2106.03611}}
  (\bibinfo{year}{2021}).
\newblock


\bibitem[Pfeiffer et~al\mbox{.}(2016)]%
        {pfeiffer2016predicting}
\bibfield{author}{\bibinfo{person}{Mark Pfeiffer}, \bibinfo{person}{Ulrich
  Schwesinger}, \bibinfo{person}{Hannes Sommer}, \bibinfo{person}{Enric
  Galceran}, {and} \bibinfo{person}{Roland Siegwart}.}
  \bibinfo{year}{2016}\natexlab{}.
\newblock \showarticletitle{Predicting actions to act predictably: Cooperative
  partial motion planning with maximum entropy models}. In
  \bibinfo{booktitle}{\emph{2016 IEEE/RSJ International Conference on
  Intelligent Robots and Systems (IROS)}}. IEEE, \bibinfo{pages}{2096--2101}.
\newblock


\bibitem[Premack and Woodruff(1978)]%
        {premack1978does}
\bibfield{author}{\bibinfo{person}{David Premack} {and} \bibinfo{person}{Guy
  Woodruff}.} \bibinfo{year}{1978}\natexlab{}.
\newblock \showarticletitle{Does the chimpanzee have a theory of mind?}
\newblock \bibinfo{journal}{\emph{Behavioral and brain sciences}}
  \bibinfo{volume}{1}, \bibinfo{number}{4} (\bibinfo{year}{1978}),
  \bibinfo{pages}{515--526}.
\newblock


\bibitem[Rae et~al\mbox{.}(2013)]%
        {rae2013influence}
\bibfield{author}{\bibinfo{person}{Irene Rae}, \bibinfo{person}{Leila
  Takayama}, {and} \bibinfo{person}{Bilge Mutlu}.}
  \bibinfo{year}{2013}\natexlab{}.
\newblock \showarticletitle{The influence of height in robot-mediated
  communication}. In \bibinfo{booktitle}{\emph{ACM/IEEE International
  Conference on Human-Robot Interaction}}. \bibinfo{pages}{1--8}.
\newblock


\bibitem[Reddy et~al\mbox{.}(2018)]%
        {reddy2018you}
\bibfield{author}{\bibinfo{person}{Sid Reddy}, \bibinfo{person}{Anca Dragan},
  {and} \bibinfo{person}{Sergey Levine}.} \bibinfo{year}{2018}\natexlab{}.
\newblock \showarticletitle{Where do you think you're going?: Inferring beliefs
  about dynamics from behavior}.
\newblock \bibinfo{journal}{\emph{Advances in Neural Information Processing
  Systems}}  \bibinfo{volume}{31} (\bibinfo{year}{2018}).
\newblock


\bibitem[Reddy et~al\mbox{.}(2020)]%
        {reddy2020assisted}
\bibfield{author}{\bibinfo{person}{Siddharth Reddy}, \bibinfo{person}{Sergey
  Levine}, {and} \bibinfo{person}{Anca~D Dragan}.}
  \bibinfo{year}{2020}\natexlab{}.
\newblock \showarticletitle{Assisted perception: optimizing observations to
  communicate state}.
\newblock \bibinfo{journal}{\emph{arXiv preprint arXiv:2008.02840}}
  (\bibinfo{year}{2020}).
\newblock


\bibitem[Sadigh et~al\mbox{.}(2016)]%
        {sadigh2016planning}
\bibfield{author}{\bibinfo{person}{Dorsa Sadigh}, \bibinfo{person}{Shankar
  Sastry}, \bibinfo{person}{Sanjit~A Seshia}, {and} \bibinfo{person}{Anca~D
  Dragan}.} \bibinfo{year}{2016}\natexlab{}.
\newblock \showarticletitle{Planning for autonomous cars that leverage effects
  on human actions.}. In \bibinfo{booktitle}{\emph{Robotics: Science and
  Systems}}.
\newblock


\bibitem[Saunderson and Nejat(2019)]%
        {saunderson2019robots}
\bibfield{author}{\bibinfo{person}{Shane Saunderson} {and}
  \bibinfo{person}{Goldie Nejat}.} \bibinfo{year}{2019}\natexlab{}.
\newblock \showarticletitle{How robots influence humans: {A} survey of
  nonverbal communication in social human--robot interaction}.
\newblock \bibinfo{journal}{\emph{International Journal of Social Robotics}}
  \bibinfo{volume}{11}, \bibinfo{number}{4} (\bibinfo{year}{2019}),
  \bibinfo{pages}{575--608}.
\newblock


\bibitem[Schaefer et~al\mbox{.}(2012)]%
        {schaefer2012beside}
\bibfield{author}{\bibinfo{person}{Sydney~Y Schaefer}, \bibinfo{person}{Iris~L
  Shelly}, {and} \bibinfo{person}{Kurt~A Thoroughman}.}
  \bibinfo{year}{2012}\natexlab{}.
\newblock \showarticletitle{Beside the point: motor adaptation without
  feedback-based error correction in task-irrelevant conditions}.
\newblock \bibinfo{journal}{\emph{Journal of Neurophysiology}}
  \bibinfo{volume}{107}, \bibinfo{number}{4} (\bibinfo{year}{2012}),
  \bibinfo{pages}{1247--1256}.
\newblock


\bibitem[Schulman et~al\mbox{.}(2017)]%
        {schulman2017proximal}
\bibfield{author}{\bibinfo{person}{John Schulman}, \bibinfo{person}{Filip
  Wolski}, \bibinfo{person}{Prafulla Dhariwal}, \bibinfo{person}{Alec Radford},
  {and} \bibinfo{person}{Oleg Klimov}.} \bibinfo{year}{2017}\natexlab{}.
\newblock \showarticletitle{Proximal policy optimization algorithms}.
\newblock \bibinfo{journal}{\emph{arXiv preprint arXiv:1707.06347}}
  (\bibinfo{year}{2017}).
\newblock


\bibitem[Schwarting et~al\mbox{.}(2019)]%
        {schwarting2019social}
\bibfield{author}{\bibinfo{person}{Wilko Schwarting}, \bibinfo{person}{Alyssa
  Pierson}, \bibinfo{person}{Javier Alonso-Mora}, \bibinfo{person}{Sertac
  Karaman}, {and} \bibinfo{person}{Daniela Rus}.}
  \bibinfo{year}{2019}\natexlab{}.
\newblock \showarticletitle{Social behavior for autonomous vehicles}.
\newblock \bibinfo{journal}{\emph{Proceedings of the National Academy of
  Sciences}} \bibinfo{volume}{116}, \bibinfo{number}{50}
  (\bibinfo{year}{2019}), \bibinfo{pages}{24972--24978}.
\newblock


\bibitem[Shi et~al\mbox{.}(2008)]%
        {shi2008performing}
\bibfield{author}{\bibinfo{person}{Lei Shi}, \bibinfo{person}{Naomi~H Feldman},
  {and} \bibinfo{person}{Thomas~L Griffiths}.} \bibinfo{year}{2008}\natexlab{}.
\newblock \showarticletitle{Performing Bayesian inference with exemplar
  models}. In \bibinfo{booktitle}{\emph{Proceedings of the Annual Meeting of
  the Cognitive Science Society}}, Vol.~\bibinfo{volume}{30}.
\newblock


\bibitem[Srivastava et~al\mbox{.}(2022)]%
        {srivastava2022assistive}
\bibfield{author}{\bibinfo{person}{Megha Srivastava}, \bibinfo{person}{Erdem
  Biyik}, \bibinfo{person}{Suvir Mirchandani}, \bibinfo{person}{Noah Goodman},
  {and} \bibinfo{person}{Dorsa Sadigh}.} \bibinfo{year}{2022}\natexlab{}.
\newblock \showarticletitle{Assistive Teaching of Motor Control Tasks to
  Humans}.
\newblock \bibinfo{journal}{\emph{arXiv preprint arXiv:2211.14003}}
  (\bibinfo{year}{2022}).
\newblock


\bibitem[Sutton and Barto(2018)]%
        {sutton2018reinforcement}
\bibfield{author}{\bibinfo{person}{Richard~S Sutton} {and}
  \bibinfo{person}{Andrew~G Barto}.} \bibinfo{year}{2018}\natexlab{}.
\newblock \bibinfo{booktitle}{\emph{Reinforcement learning: An introduction}}.
\newblock \bibinfo{publisher}{MIT press}.
\newblock


\bibitem[Tian et~al\mbox{.}(2022)]%
        {tian2022safety}
\bibfield{author}{\bibinfo{person}{Ran Tian}, \bibinfo{person}{Liting Sun},
  \bibinfo{person}{Andrea Bajcsy}, \bibinfo{person}{Masayoshi Tomizuka}, {and}
  \bibinfo{person}{Anca~D Dragan}.} \bibinfo{year}{2022}\natexlab{}.
\newblock \showarticletitle{Safety assurances for human-robot interaction via
  confidence-aware game-theoretic human models}. In
  \bibinfo{booktitle}{\emph{2022 International Conference on Robotics and
  Automation (ICRA)}}. IEEE, \bibinfo{pages}{11229--11235}.
\newblock


\bibitem[Tierney and Kadane(1986)]%
        {tierney1986accurate}
\bibfield{author}{\bibinfo{person}{Luke Tierney} {and}
  \bibinfo{person}{Joseph~B Kadane}.} \bibinfo{year}{1986}\natexlab{}.
\newblock \showarticletitle{Accurate approximations for posterior moments and
  marginal densities}.
\newblock \bibinfo{journal}{\emph{Journal of the american statistical
  association}} \bibinfo{volume}{81}, \bibinfo{number}{393}
  (\bibinfo{year}{1986}), \bibinfo{pages}{82--86}.
\newblock


\bibitem[Ullman and Tenenbaum(2020)]%
        {annurev-devpsych-121318-084833}
\bibfield{author}{\bibinfo{person}{Tomer~D. Ullman} {and}
  \bibinfo{person}{Joshua~B. Tenenbaum}.} \bibinfo{year}{2020}\natexlab{}.
\newblock \showarticletitle{Bayesian Models of Conceptual Development: Learning
  as Building Models of the World}.
\newblock \bibinfo{journal}{\emph{Annual Review of Developmental Psychology}}
  \bibinfo{volume}{2}, \bibinfo{number}{1} (\bibinfo{year}{2020}),
  \bibinfo{pages}{533--558}.
\newblock
\urldef\tempurl%
\url{https://doi.org/10.1146/annurev-devpsych-121318-084833}
\showDOI{\tempurl}
\showeprint{https://doi.org/10.1146/annurev-devpsych-121318-084833}


\bibitem[Vaswani et~al\mbox{.}(2017)]%
        {vaswani2017attention}
\bibfield{author}{\bibinfo{person}{Ashish Vaswani}, \bibinfo{person}{Noam
  Shazeer}, \bibinfo{person}{Niki Parmar}, \bibinfo{person}{Jakob Uszkoreit},
  \bibinfo{person}{Llion Jones}, \bibinfo{person}{Aidan~N Gomez},
  \bibinfo{person}{{\L}ukasz Kaiser}, {and} \bibinfo{person}{Illia
  Polosukhin}.} \bibinfo{year}{2017}\natexlab{}.
\newblock \showarticletitle{Attention is all you need}.
\newblock \bibinfo{journal}{\emph{Advances in neural information processing
  systems}}  \bibinfo{volume}{30} (\bibinfo{year}{2017}).
\newblock


\bibitem[Waugh et~al\mbox{.}(2010)]%
        {waugh2010inverse}
\bibfield{author}{\bibinfo{person}{Kevin Waugh}, \bibinfo{person}{Brian~D
  Ziebart}, {and} \bibinfo{person}{J~Andrew Bagnell}.}
  \bibinfo{year}{2010}\natexlab{}.
\newblock \showarticletitle{Inverse Correlated Equilibrium for Matrix Games}.
\newblock \bibinfo{journal}{\emph{Advances in Neural Information Processing
  Systems}} (\bibinfo{year}{2010}).
\newblock


\bibitem[Weiss et~al\mbox{.}(2002)]%
        {weiss2002motion}
\bibfield{author}{\bibinfo{person}{Yair Weiss}, \bibinfo{person}{Eero~P
  Simoncelli}, {and} \bibinfo{person}{Edward~H Adelson}.}
  \bibinfo{year}{2002}\natexlab{}.
\newblock \showarticletitle{Motion illusions as optimal percepts}.
\newblock \bibinfo{journal}{\emph{Nature neuroscience}} \bibinfo{volume}{5},
  \bibinfo{number}{6} (\bibinfo{year}{2002}), \bibinfo{pages}{598--604}.
\newblock


\bibitem[Xie et~al\mbox{.}(2020)]%
        {xie2020learning}
\bibfield{author}{\bibinfo{person}{Annie Xie}, \bibinfo{person}{Dylan~P Losey},
  \bibinfo{person}{Ryan Tolsma}, \bibinfo{person}{Chelsea Finn}, {and}
  \bibinfo{person}{Dorsa Sadigh}.} \bibinfo{year}{2020}\natexlab{}.
\newblock \showarticletitle{Learning latent representations to influence
  multi-agent interaction}.
\newblock \bibinfo{journal}{\emph{Conference on Robot Learning}}
  (\bibinfo{year}{2020}).
\newblock


\bibitem[Ziebart et~al\mbox{.}(2008)]%
        {ziebart2008maximum}
\bibfield{author}{\bibinfo{person}{Brian~D Ziebart}, \bibinfo{person}{Andrew~L
  Maas}, \bibinfo{person}{J~Andrew Bagnell}, {and} \bibinfo{person}{Anind~K
  Dey}.} \bibinfo{year}{2008}\natexlab{}.
\newblock \showarticletitle{Maximum entropy inverse reinforcement learning.}.
  In \bibinfo{booktitle}{\emph{Aaai}}, Vol.~\bibinfo{volume}{8}. Chicago, IL,
  USA, \bibinfo{pages}{1433--1438}.
\newblock


\end{thebibliography}

\clearpage 

\appendix 

\section{Appendix}
\subsection{Derivation: Gaussian integral under LQ approximation}
\label{app:gaussian_integral}

Here we derive the closed-form solution to the denominator from \eqref{eq:human_policy} under the LQ-approximation. First, we recall the Gaussian integral:

\bigskip 
\noindent \textit{\textbf{Theorem: Gaussian Integral \cite{tierney1986accurate}.}  Let $M \in \mathbb{R}^{n \times n}$ be a symmetric, positive-definite matrix and $x \in \mathbb{R}^n$. Then:}
\begin{equation}
    \int \exp\Big( -\frac{1}{2} x^\top M x + b^\top x\Big) d^n x = \sqrt{\frac{(2\pi)^n}{\det(M)}} \exp \Big(\frac{1}{2} b^\top M^{-1} b \Big).
\end{equation}

\bigskip 
\noindent \textit{\textbf{Theorem: Infinite-horizon Linear-Quadratic Regulator \cite{bertsekas2011dynamic}.} Let the discrete-time dynamics be linear, $x^{t+1} = Ax^t + Bu^t$, and the cost quadratic, $x^\top Q x + u^\top R u$. Then the infinite-horizon optimal cost-to-go $J$ and optimal control $u^*(x)$ are:}
\begin{align}
    J(x) &= x^\top P x \\
    u^*(x) &= -K x
\end{align}
\textit{where $P$ is the unique, positive-definite fixed point of the infinite-horizon, discrete-time Ricatti equation (DARE):}
\begin{align}
P &= A^{\top}PA - A^{\top}PB(R+B^{\top}PB)^{-1}B^{\top}PA + Q
\label{eq:appendix_dare}
\end{align}
\textit{and the feedback matrix $K = (R + B^\top P B)^{-1} B^\top P A$.}

\bigskip 
\noindent\textbf{Derivation.} Let that the physical dynamics be linear and the reward is quadratic in state and control. Assume that we have approximated the state-action $Q_\mathrm{H}$ function as:
\begin{equation}
   Q_\mathrm{H}(\jointstate, u) = - x^\top Q x - u^\top R u -(\jointstate')^\top P (\jointstate')
   \label{eq:appendix_q_value}
\end{equation}
where $x' = Ax + Bu$ is the next state and $P$ is the solution to \eqref{eq:appendix_dare}.
Plugging in \eqref{eq:appendix_q_value} into the denominator of the human policy, we obtain:
\begin{align}
    \int &\exp\Big(Q_\mathrm{H}(x, u) \Big) du \\
    &= \int \exp\Big(-x^\top Q x - u^\top R u - (Ax + Bu)^\top P (Ax + Bu) \Big) du \\
    &= \exp\Big(-x^\top Q x - x^\top A^\top P A x\Big) \cdot \\ & \quad \quad \int \exp\Big(-\frac{1}{2} u^\top (2R + 2B^\top P B)u + (- 2x^\top A^\top P B) u \Big) du
\end{align}
We see that if we let $M := 2R + 2B^\top P B$ and $b := -2x^\top A^\top P B$ then we can directly take the Gaussian integral and obtain:
\begin{align}
    &= \exp\Big(-x^\top Q x - x^\top A^\top P A x\Big) \sqrt{\frac{(2\pi)^{m}}{\det(2R + 2B^\top P B)}} \cdot \\ &\quad \quad \quad \exp \Big( (-2x^\top A^\top P B)^\top [2R + 2B^\top P B]^{-1} (-2x^\top A^\top P B)\Big)\\
    &= \exp\Big(x^\top \big[ A^\top P B (R + B^\top P B)^{-1} B^\top P A - Q - A^\top P A\big] x \Big) \cdot \\ & \quad \quad \quad \sqrt{\frac{(2\pi)^{m}}{\det(2R + 2B^\top P B)}}. 
\end{align}
Interestingly, we see that the exponent contains the (negated) DARE equation from \eqref{eq:appendix_dare} within the brackets. Substituting $P$ back in, we obtain:
\begin{align}
    \int &\exp\Big(Q_\mathrm{H}(x, u) \Big) du  = \exp\Big(- x^\top P x \Big) \sqrt{\frac{(2\pi)^{m}}{\det(2R + 2B^\top P B)}}. 
\end{align}
We can further simplify this equation by 
\begin{align}
    x^\top P x &= \min_u \Big[ x^\top Q x + u^\top R u + (Ax + Bu)^\top P (Ax + Bu) \Big] \\ 
    &= x^\top Q x + (u^*)^\top R (u^*) + (Ax + Bu^*)^\top P (Ax + Bu^*) \\
    &= Q_\mathrm{H}(x,u^*)
\end{align}
where $u^*$ is the optimal control at state $x$. Thus, we can obtain our final simplified form:
\begin{equation*}
    \int \exp\Big(Q_\mathrm{H}(x, u) \Big) du  = \exp\Big(- Q_\mathrm{H}(x,u^*) \Big) \sqrt{\frac{(2\pi)^{m}}{\det(2R + 2B^\top P B)}}. 
\end{equation*}

\begin{figure}[h]
    \centering
    \includegraphics[width=0.4\textwidth]{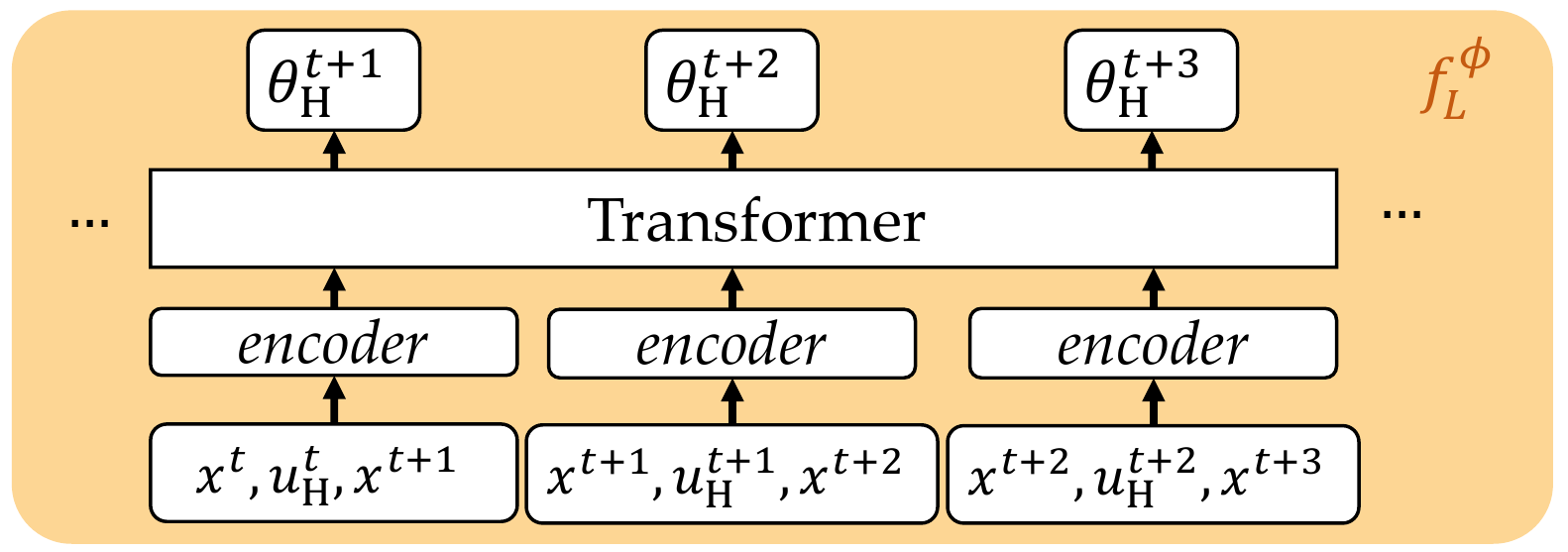}
    \caption{Architecture of transformer representing $\approxparamdyn$.}
    \label{fig:transformer_model}
\end{figure}

\subsection{Details on a gradient-based solution to inferring the dynamics of human learning}
\label{app:gradient_based_soln}
To optimize the transformer-based model of human learning dynamics, we need to compute the gradient of our inference objective (in Equation~\eqref{eq:learning_problem} and referred to here as $\mathcal{L}$) with respect to the neural network parameters, $\phi$:
$$
\frac{\partial\mathcal{L} }{\partial \phi} = \frac{\partial\mathcal{L} }{\partial \mathbb{\param}}\frac{\partial\mathbf{\param} }{\partial \phi}.
$$ 
The second component, $\frac{\partial\mathbf{\param} }{\partial \phi}$, is the gradient of the transformer's internal model predictions with respect to the neural network weights and is readily available since the transformer is differentiable. 
However, the first component 
$$
\frac{\partial\mathcal{L} }{\partial \mathbf{\param}}= \big[\frac{\partial \log \mathbb{P}(\uH^t \mid  \jointstate^t;\param^t) }{\partial \mathbf{\param}^t}\big], 
$$ 
which is the human's policy gradient with respect to the human's internal model parameter, is a key challenge. 
This is because the human's policy $\mathbb{P}(\uH \mid \jointstate, \param)$ depends on $P_{\param}$ through the $Q_\mathrm{H}$-value. 
Recall that $P_{\param}$ is the solution to DARE in Equation~\eqref{eq:DARE} which depends on the matrices $A, B, Q, R$ and $P_{\param}$. 
Regardless of if the human's internal model parameter is the physical dynamics $\param := (A, B)$ or the reward weights $\param := (Q, R)$, the human's policy gradient requires differentiating through the DARE function, which is non-obvious. Leveraging recent work \cite{east2020infinite} that treats the DARE as an implicit function of $(A, B, Q, R)$, we obtain closed-form Jacobians
$\frac{\partial P }{ \partial A}$, $\frac{\partial P }{ \partial B}$, $\frac{\partial P }{ \partial Q}$, and $\frac{\partial P }{ \partial R}$. The precise form of these can be found in Proposition 2 of \cite{east2020infinite}.
Thus, we can efficiently compute $\frac{\partial\mathcal{L} }{\partial \phi}$ and infer the human's learning dynamics via gradient-based optimization. 

\subsection{Training the dynamics model of human learning}
\label{app:training_details}

To enhance the reproducibility of inferring $\approxparamdyn$, we present the architecture and optimization details here. 
The encoder for encoding $(x^t, \uH^t, x^{t+1})$ is a multilayer perceptron with 3 fully-connected layers in all settings. We use the Hugging Face's implementation \cite{hugging2022transformer} of the transformer encoder \cite{vaswani2017attention} to represent the human's learning dynamics, and use the Adam optimizer to train the neural network.

In both the simulated experiments and in the user study, we use the same transformer architecture to represent the dynamics of human learning, with only the output layer size adjusted per each task to appropriately model the human's internal model $\param$. 
From Section~\ref{subsec:teaching_physics_dyn}, in \textbf{Lunar Lander} the output size is 2-dimensional, representing the $B$-vector that the human is estimating and in \textbf{Robot Arm Teleoperation} the output is 4-dimensional to account for the diagonal elements of $B$ and $w$. 
From Section~\ref{subsec:reward_manipulation}, in \textbf{Goal Influence} the output size is 2-dimensional, representing the probability (i.e. human belief) over the first and the second tray goals (the probability over the third goal is implicitly defined as one minus the probability of the other two goals combined), while in \textbf{Preference Influence} the output size is 3-dimensional to represent the diagonal terms of the $Q \in \mathbb{R}^{3 \times 3}$. 
Finally, in the user study from Section~\ref{sec:user_study},  the output is 4-dimensional to account for the diagonal elements of $B$ and $w$.
Note that the human's initial internal model ($\param^0$) is implicitly estimated at the beginning of the input when predicting $\param^1$.

\subsection{User Study: Human and Robot Action Alignment}
\label{app:user_study_alignment}
\begin{figure}[h]
    \centering
    \includegraphics[width=0.5\textwidth]{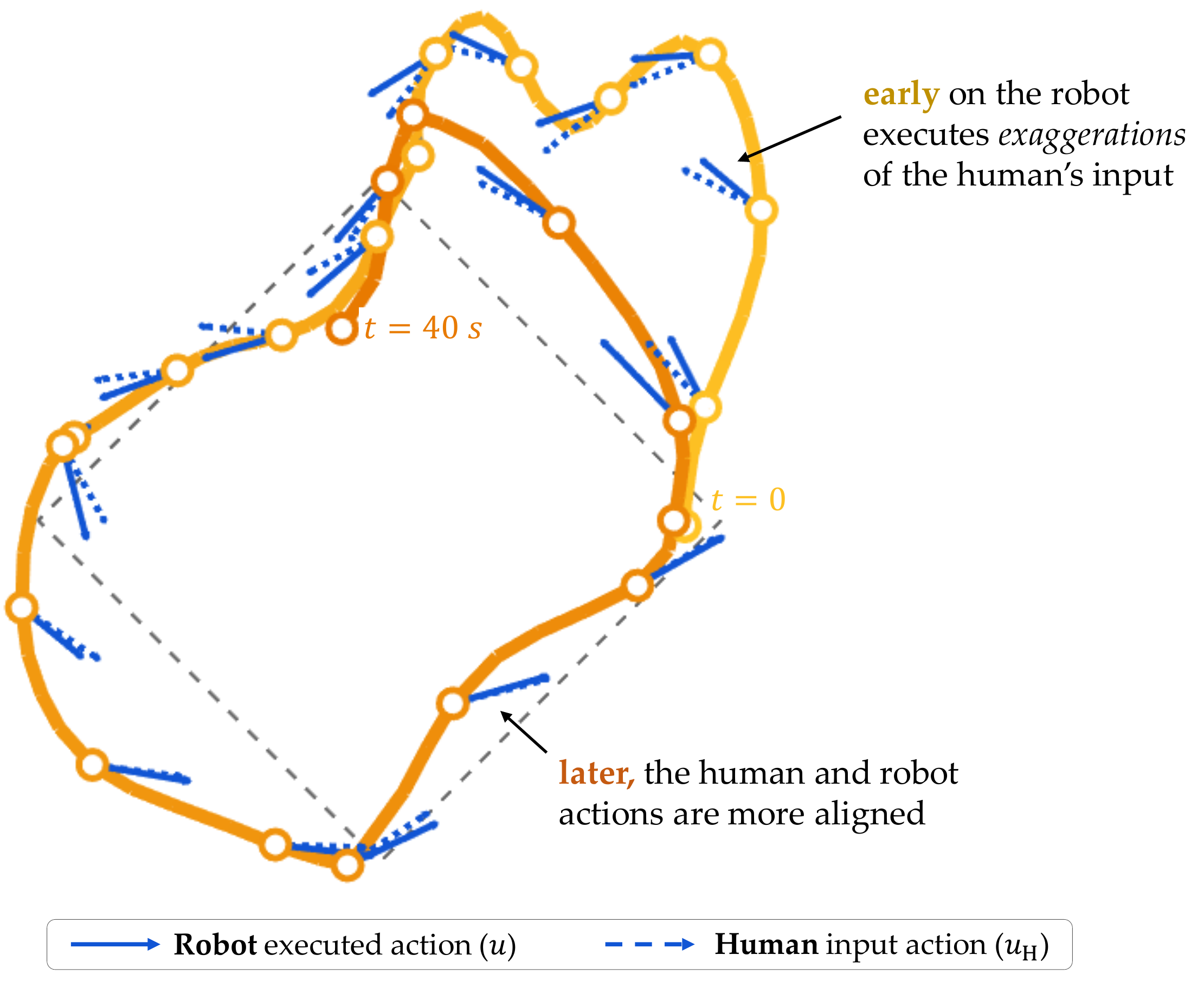}
    \caption{An example participant trajectory from Section~\ref{sec:user_study}. Blue vectors show the executed robot actions (in solid line) and human input action (in dashed line) at timesteps sampled at 0.5 s.}
    \label{fig:user_study_uH_uR_alignment}
\end{figure}

We looked at the user study data and investigated how the human input actions compared to the executed robot actions under our teaching method. Recall that the robot executes actions according to \eqref{eq:shared_control}: $u = \alpha \cdot \uR + (1 - \alpha) \cdot \uH$ where $\alpha = 0.5$ for the duration of our user study and $\uR$ is generated according to our influence-aware planning method. In Fig.~\ref{fig:user_study_uH_uR_alignment} we plot a sample participant trajectory and the robot executed actions (solid blue vector) and human input actions (dashed blue vector) at 1.5 s time intervals. Qualitatively, we see that early on the human and robot's actions are misaligned. 
Intuitively, since the robot's planning objective is to quickly align the human's mental model of the physics with the robot's physics model, the robot plans to execute an exaggeration of the human's input in hopes of quickly changing their mind.
Later on, we see that the human and robot actions become more aligned as the human learns to be a better teleoperator. 

\end{document}